\documentclass{article}

\usepackage{PRIMEarxiv}

\usepackage[utf8]{inputenc} 
\usepackage[T1]{fontenc}    
\usepackage{hyperref}       
\usepackage{url}            
\usepackage{booktabs}       
\usepackage{amsfonts}       
\usepackage{nicefrac}       
\usepackage{microtype}      
\usepackage{lipsum}
\usepackage{fancyhdr}       
\usepackage{graphicx}       
\graphicspath{{media/}}     

\usepackage[numbers]{natbib}

\usepackage{multirow}
\usepackage{float}
\usepackage{subfig} 
\usepackage{amsmath}
\usepackage{mathtools}
\usepackage{bm}
\usepackage{multirow}
\usepackage{enumitem}
\usepackage[ruled,vlined,linesnumbered,noend]{algorithm2e}
\usepackage[table,xcdraw]{xcolor}

\usepackage{combelow}

\usepackage{pgfplots}
\usetikzlibrary{patterns,arrows}
\pgfplotsset{compat=1.17}

\newcommand{\resultlowclr}{\cellcolor[HTML]{FEF9C3}}
\newcommand{\resultmidclr}{\cellcolor[HTML]{FEF08A}}
\newcommand{\resulthighclr}{\cellcolor[HTML]{FDE047}}

\title{ATESA-B{\AE}RT: A Heterogeneous Ensemble Learning Model for Aspect-Based Sentiment Analysis}

\author{
Elena-Simona Apostol, Alin-Georgian Pisic\u{a}, and Ciprian-Octavian Truic\u{a} \\
University Politehnica of Bucharest, Bucharest, Romania \\
\texttt{elena.apostol@upb.ro, alin.pisica@stud.acs.upb.ro, ciprian.truica@upb.ro}
}

\begin{document}
\maketitle

\begin{abstract}
The increasing volume of online reviews has made possible the development of sentiment analysis models for determining the opinion of customers regarding different products and services.
Until now, sentiment analysis has proven to be an effective tool for determining the overall polarity of reviews.
To improve the granularity at the aspect level for a better understanding of the service or product, the task of aspect-based sentiment analysis aims to first identify aspects and then determine the user's opinion about them.
The complexity of this task lies in the fact that the same review can present multiple aspects, each with its own polarity.
Current solutions have poor performance on such data.
We address this problem by proposing ATESA-B{\AE}RT, a heterogeneous ensemble learning model for Aspect-Based Sentiment Analysis.
Firstly, we divide our problem into two sub-tasks, i.e., Aspect Term Extraction and Aspect Term Sentiment Analysis.
Secondly, we use the \textit{argmax} multi-class classification on six transformers-based learners for each sub-task.
Initial experiments on two datasets prove that ATESA-B{\AE}RT outperforms current state-of-the-art solutions while solving the many aspects problem.
\end{abstract}


\section{Introduction}

Sentiment Analysis is currently an efficient and wildly used tool for extracting the overall user opinion from product and service reviews~\cite{Do2019}.
Aspect-Based Sentiment Analysis (ABSA) is an extension of Sentiment Analysis that introduces an additional granularity level represented by aspect identification and polarity prediction.
The current literature proposes various models, from simple classifiers that solve specific tasks (aspect extraction, keywords extraction, or sentiment analysis) to more complex architectures involving transformers, pre-trained models, or chained neural networks.

Even though the general sentence and document level sentiment analysis has been used in a many domains~\cite{Mitroi2020,Petrescu2019,Petrescu2023,Truica2021}, the ABSA task is not something that is implemented very often due to the huge challenge created by the large variety of possible aspects and categories.
ABSA  aims to identify the fine-grained polarity towards specific aspects of the text, allowing users to easily visualize and understand complex language formations, giving a granular understanding of the quality and the transmitted polarities.

Over the last few years, the pre-trained language models have proven to be effective in replacing and improving the feature engineering phase of the development.
Transformers like BERT~\cite{Devlin2019} (Bidirectional Encoder Representations from Transformers) and GPT~\cite{Radford2018} (Generative Pre-trained Transformer) have introduced the possibility of creating state-of-the-art NLP models with as little as a fine-tuning process.
However, due to the complexity of the ABSA task, simple fine-tuning is not enough for obtaining an accurate solution.

The main motivation for this paper is to evaluate the performance of different fine-tuned transformer-based models for the task of ABSA in order to propose a novel ensemble learning model, i.e.,~ATESA-B{\AE}RT.
Ensemble Learning is a type of machine learning technique that can improve overall performance by integrating multiple models and applying specific strategies based on group decision-making.
To reduce the difficulty of the ABSA problem, we propose its division into two sub-tasks: 
(1) Aspect Term Extraction (ATE), in which we aim to identify the correct aspects of each input, and
(2) Aspect Term Sentiment Analysis (ATSA), which obtains the polarity, either negative, neutral, or positive, for each aspect of the review.
For training and testing, we use two real-world English online review datasets consisting of $\sim2\,000$ and $\sim4\,300$ sentences, respectively, with one aspect, multiple or no aspects.

The research questions for our study are as follows:
\begin{itemize}
    \item[(\textit{Q1})] Can we improve the extraction and detection of aspect terms by using a heterogeneous ensemble model?
    \item[(\textit{Q2})] Can there be a balance between accuracy and runtime required for fine-tuning and training a transformer for the ABSA task?
\end{itemize}

Our contributions are as follows:
\begin{itemize}
    \item[(\textit{C1})] We propose ATESA-B{\AE}RT, a novel heterogeneous ensemble learning model that aims to identify the fine-grained polarity towards specific aspects of a text by employing different deep learning models and multiple fine-tuned transformers.
    \item[(\textit{C2})] We propose the division of the ABSA task into two sub-tasks, i.e., ATE and ATSA, and performed a detailed analysis using multiple transformers and deep learning models and compared the results with the state-of-the-art solutions.
    \item[(\textit{C3})] We trained and tested our models on two real-world datasets and performed an in-depth analysis of the obtained results.
\end{itemize}

This article is structured as follows.
Section~\ref{sec:related_work} discusses the current research related to ABSA.
Section~\ref{sec:methodology} introduces the proposed ensemble architecture and the employed transformers as well as the neural network models used for classification.
Section~\ref{sec:experimental_results} presents the datasets and analyzes the experimental results.
The section ends with discussions regarding our findings and hints at currently identified challenges.
Section~\ref{sec:conclusions} presents the conclusions and outlines future directions.

\section{Related Work}\label{sec:related_work}

In this section, we present and analyze the current state-of-the-art solutions for aspect extraction and ABSA.

\paragraph{\textbf{Aspect and Feature Extraction.}}
Features or aspects are entities present in the text on which the focus and sentiments are targeted.
The main goal of the aspect extraction is to identify the targeted entities (e.g., products, topics, places) or groups of entities and match them together with the classified polarity, in order to identify the positive and negative areas and how the overall feelings of the reviewer can be improved.

\citet{Xu2018} propose an aspect extraction method based on two embedding layers (one for general-purpose and one domain-specific) combined with four convolutional neural networks.
\citet{he-etal-2017-unsupervised} present a neural approach in aspect extraction, focused on discovering coherent aspects.
The model exploits the distribution of word co-occurrences using word embeddings.
\citet{Pradhan2021} use for aspect detection a frequency-based method, by calculating terms' frequency using unigrams and then identifying aspects with a bigram approach.
\citet{Phan2020} propose an aspect extraction architecture based on RoBERTa transformer~\cite{Liu2019} and POS and dependency-based embeddings. 

\paragraph{\textbf{Aspect-Based Sentiment Analysis}}
ABSA aims to identify and pair the aspect extracted with the corresponding sentiment by employing both lexicon based and machine learning based techniques.
Current solutions use machine learning based techniques to extract aspects and sentiments.
\citet{Tang2016} present an approach to handle the aspect identification and polarity matching based on Long Short Term Memory (LSTM) networks.
\citet{absausingbert} solves the ABSA problem by creating 3 models that use BERT for sentence pair classification, i.e., one for aspect classification, one for polarity classification and one that combines the previous two for obtaining the final result.
\citet{Marcacini2021} make use of DeBERTa (Decoding-enhanced BERT with Disentangled Attention)~\cite{PengchengHe2021}, which is an improvement of BERT and RoBERTa transformers by using a disentangled attention mechanism and an enhanced mask decoder.

Recent works also use graph neural networks to model connections between aspects and opinion words.
Sentic GCN~\cite{Liang2022} is a graph convolutional network ABSA solution based on SenticNet that takes account of both the dependencies of contextual words and aspect words and the affective information between opinion words and the aspect.
DualGCN~\cite{Li2021} is a dual graph convolutional networks model for the ABSA task that considers the semantic correlation and syntactic structure within a sentence.

Some solutions use transfer learning to tackle the problem of not having large amount of labeled data to train the ABSA model.
\citet{Tao2020} propose such a solution by extending the ABSA methods with multi-label classification capabilities.
Their transfer learning models use BERT and XLNet~\cite{Yang2019}.
Another transfer learning-based approach is proposed by \citet{Bensoltane2021}. Their solution is built based on the Arabic version of the BERT model, i.e., AraBERT.

\section{Methodology}\label{sec:methodology}

\begin{figure*}[!htbp]
\centering
\includegraphics[width=1\textwidth]{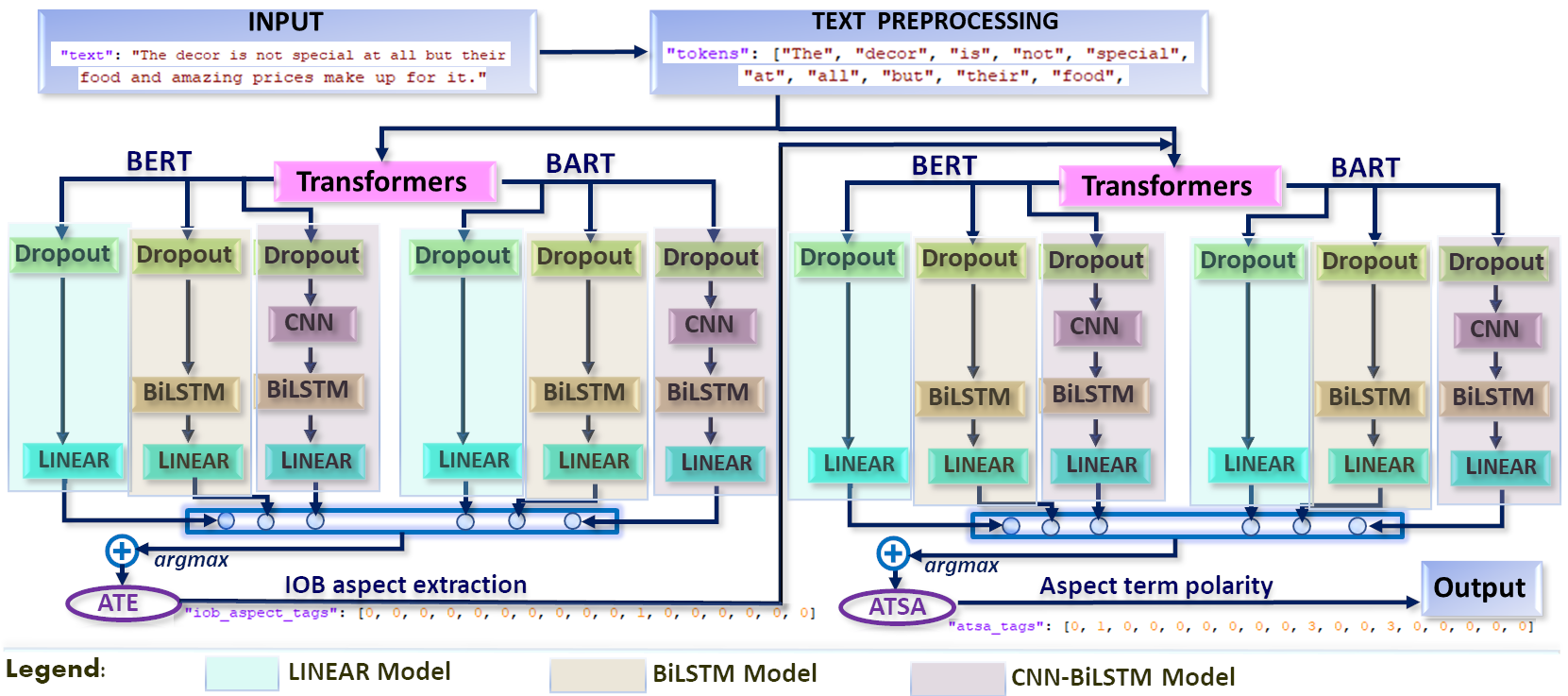}
\caption{ATESA-B{\AE}RT Ensemble Architecture}
\label{fig:architecture}
\end{figure*}

In this section, we present the overall ensemble architecture while also detailing the tagging employed to extract aspect terms, the proposed neural models and their implementation, as well as the web application we developed for interacting with our proposed solution.
Figure~\ref{fig:architecture} presents the workflow of our novel architecture ATESA-B{\AE}RT: Aspect Term Extraction and Sentiment Analysis using a heterogeneous ensemble learning model with BERT and BART.

In our ensemble architecture, we have two main branches:
\begin{itemize}
    \item[(\textit{1})] The Aspect Term Extraction (ATE) branch, where we train an ensemble consisting of six different models to identify the aspects of each pre-processed input text, and
    \item[(\textit{2})] The Aspect Term Sentiment Analysis (ATSE) branch, where we train an ensemble consisting of six different models to obtain the polarity, either negative, neutral or positive, for each discovered aspect by the ATE branch.
\end{itemize}

\subsection{Text preprocessing}
We use the same preprocessing techniques for both branches.
To prepare the text for the neural networks and to better determine the polarity of each aspect category and term, we use tokenization.
Thus, we split the text into words.
We apply no other type of preprocessing, as we want to capture the context and the language inflections as much as possible.
For this step, we employ \href{https://spacy.io/}{spaCy}~\cite{SpaCy}, an NLP framework that obtains state-of-the-art results on NLP tasks of part of speech identification and sentence splitting over English datasets.

\subsection{Aspect term extraction}

\begin{figure}[!htbp]
\centering
\includegraphics[width=0.75\columnwidth]{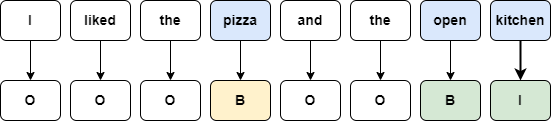}
\caption{IOB Tagging}
\label{fig:iobtaggingexample}
\end{figure}

The target of the ATE branch is to determine for each sentence the IOB (Inside-Outside-Beginning) tags that mark the aspect terms of the reviews.
IOB tagging~\cite{Ramshaw1995} is a technique used for tagging and extracting tokens, mainly used for name entity recognition~\cite{Cho2013}.
Each important token receives a corresponding value from \textbf{I} (inside), \textbf{O} (outside), or \textbf{B} (beginning).
The B value represents the beginning of the tag, the elements are marked with \textbf{I} if they are inside of an important attribute / entity / aspect term, and the \textbf{O} value is assigned to all the outside elements.
In Figure~\ref{fig:iobtaggingexample}, we present how to tag the sentence "I liked the pizza and the open kitchen", where the aspect terms identified being "pizza" and "open kitchen".
Since "pizza" is a single noun, it will be marked with the \textbf{B} tag.
While the "open kitchen", being compound, will be identified using both the \textbf{B} and \textbf{I} tags.

\subsection{Neural Models employed in the Ensemble}

Each of the two main branches, i.e., ATE and ATSA, is using the same type of Ensemble.
The ATE branch predicts the IOB tags that mark the aspect, while the ATSA branch predicts the polarity of each extracted aspect.
The main deep learning models considered for each branch of the proposed Ensemble are:
(1) \textsc{Linear},
(2) \textsc{BiLSTM}, and
(3) \textsc{CNN-BiLSTM}.
The models receive as input the preprocessed textual data, which is in turn used to extract word embeddings.
We use both pre-trained and fine-tuned BERT~\cite{Devlin2019} and BART~\citep{Lewis2020} word embeddings, but the models are generic enough to accept any type of word embeddings.
As small and medium length texts can lead the neural network to overfitting, we employ a dropout layer after the Transformer layer as a regularization method to improve the generalization and reduce the overfitting.

The \textsc{Linear} model (Figure~\ref{fig:architecture} - light blue cell) contains the following layers: (1) transformer, (2) dropout, and (3) linear layer.
The dropout layer is used to balance the input data and classes and reduce the overfitting scenario.
The linear layer is used to predict the IOB tags for the ATE Branch or the aspect terms polarity of the ATSA Branch.
In order to process the data, the transformer has to receive the tokenized text and then assigns to each token an embedding.

The \textsc{Linear} model is also used to fine-tune BERT and BART.
The fine-tuned version of BERT and BART, respectively are the saved transformer models obtained after the first training phase.
These transformer models are in turn used as the fine-tuned version for all the models.
Once fine-tuned from a general version of the model, we obtain a specialized model over the dataset.
As a downside of creating the fine-tuned transformer models, we mention: 
(1) additional preprocessing time for producing the specialized word embedding, and
(2) additional disk space for storing the model.

The \textsc{BiLSTM} model (Figure~\ref{fig:architecture} - light orange cell) uses an additional Bidirectional LSTM (BiLSTM) layer between the dropout and linear layer to better understand and analyze the contextual dependencies between the tokens.
We use a BiLSTM instead of an LSTM (Long short-term memory) because the BiLSTM improves on the traditional LSTM in terms of scoring and efficiency~\cite{Augustyniak2021}.
Using the \textsc{BiLSTM} model, we aim to analyze and understand how contextual dependencies influence polarity prediction.
Thus, using the BiLSTM layer, we create and generalize the connections between words and the memorized patterns over multiple reviews applied to the same domain and aspect terms.

The \textsc{CNN-BiLSTM} model (Figure~\ref{fig:architecture} - light purple cell) uses a 1-dimension convolutional (Conv1D) layer before the BiLSTM to obtain a feature map of the review.
Since the number of trainable parameters needed for a convolutional network is of moderate size, we expect the training time to not be affected in a substantial manner.
The convolutional layer manages to improve the grouped textual analysis by executing kernel convolutions over multiple tokens at a time.
Furthermore, this layer also improves the contextual connections between the words.
Thus, combining the BiLSTM with the convolution operations in the hidden layers might improve the prediction task.

Regardless of the model, the final layer is a linear fully connected layer.
For the aspect term extraction procedure, this layer contains 3 units that predict the possible labels for the IOB tagging method as follows: 0 for the Outside tag, 1 for the Beginning tag, and 2 for the Inside tag.
For the sentiment analysis task, the final layer contains 4 units that predict the sentiment as follows: 0 if no sentiment is extracted, 1 if the sentiment is negative, 2 for neutral, and 3 for positive.
The proposed Ensembles are using an \textit{argmax} multi-class classification method.
The outputs of each of the six proposed models are fed the \textit{argmax} function.

\subsection{Web Application}~\label{sec:webapplication}

\begin{figure}[!htbp]
\centering
\includegraphics[width=0.75\columnwidth]{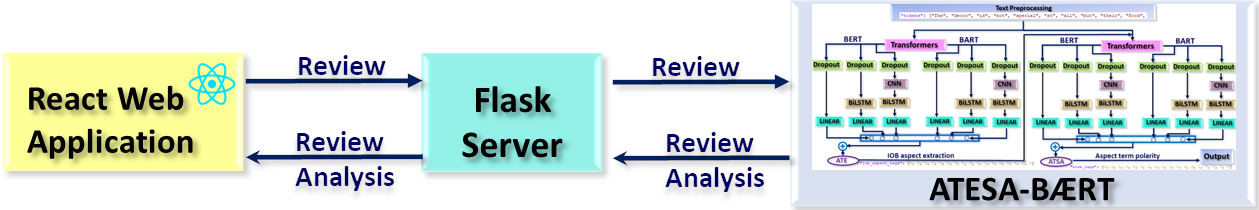}
\caption{System's data flow: Web application architecture}
\label{fig:systemarchitecture}
\end{figure}

We also developed a web application to easily test, analyze, and visualize the behavior of the ensemble models on different inputs, scenarios, and domains.
The web application is developed in React and communicates through a REST API with a Flask server that processes the data received from the front-end and forwards it to the ensemble models (Figure~\ref{fig:systemarchitecture}).

\begin{figure}[!htbp]
    \centering
    \captionsetup{justification=centering}
    \subfloat[Input custom text]{%
        \includegraphics[width=0.49\columnwidth]{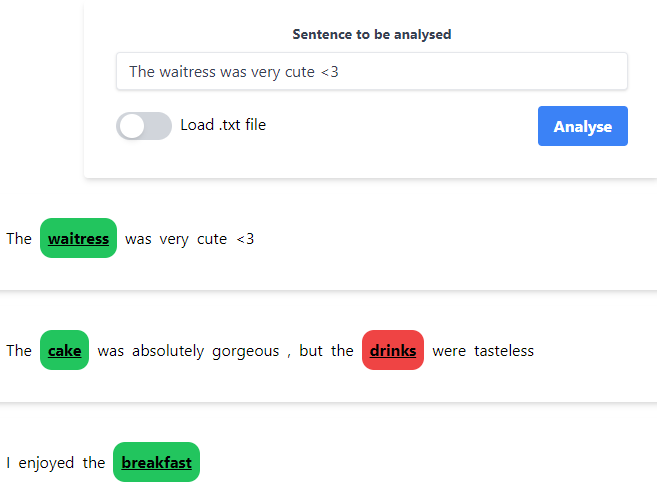}
        \label{fig:webappcustominput}%
        }%
    \hfill%
    \subfloat[Input file with multiple reviews]{%
        \includegraphics[width=0.49\columnwidth]{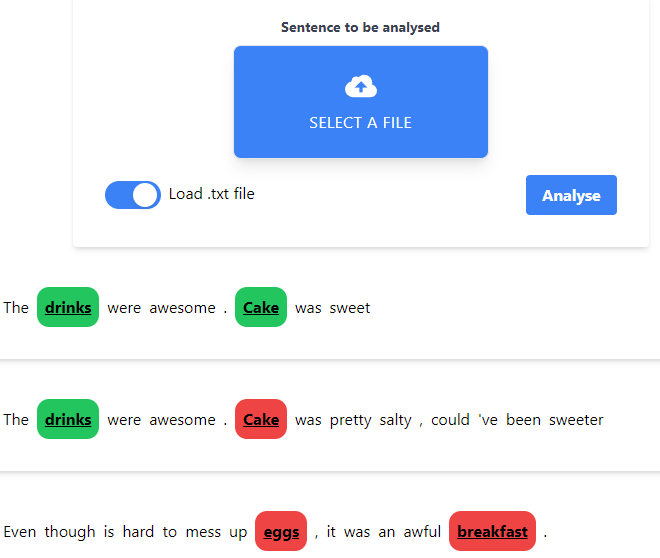}
        \label{fig:webappfileinput}%
        }%
    \caption{Web application}
    \label{fig:webapp_ui}
\end{figure}

When using the web application, the user can either
(1) input a custom text to be analyzed (Figure~\ref{fig:webappcustominput}), or
(2) upload a text file with multiple reviews, containing one review per line (Figure~\ref{fig:webappfileinput}).
The reviews are then forwarded to the backend which returns the processed data in the form of JSON objects.
To create an easy to follow processing flow, during the upload of a file containing multiple reviews, the results will appear one at a time as the model finishes processing them.
The identified aspect terms are marked with a bold, underlined font, and the sentiments extracted are highlighted using a green background for positive sentiments, red for negative, and yellow for neutral polarities.

\subsection{Implementation}~\label{sec:implementation}

We implement the models using the \href{https://pytorch.org/}{PyTorch} framework.
PyTorch is a machine learning library~\cite{Paszke2019} that gives full control to the developer over the designing of the model and offers support for CPU and GPU processing.

The backend of our web application is implemented in \href{https://flask.palletsprojects.com/en/2.1.x/}{Flask}, a web framework used for creating simple and accessible APIs.
The web application is written in \href{https://reactjs.org/}{React}, a web framework that structures a single page application (SPA) in modular and reusable components.
We use \href{https://tailwindcss.com/}{Tailwind CSS}, a utility-first CSS framework that contains atomic classes that can directly be applied, inline, to HTML components, use for the styling.
In this way, we ensure an easily scalable and upgradable interface by keeping a structured method for creating new elements and components.

The textual data preprocessing is done as follows:
(1) the whole text is tokenized, a procedure in which each word is assigned a unique identifier (token);
(2) the tokenized sentences are then batched in sequences of multiple inputs together, in order to advance to the training / testing steps;
(3) the obtained batches are then gathered together into a data-loader that handles the input serving towards the model.

\begin{figure}[!htbp]
\centering
\includegraphics[width=0.75\columnwidth]{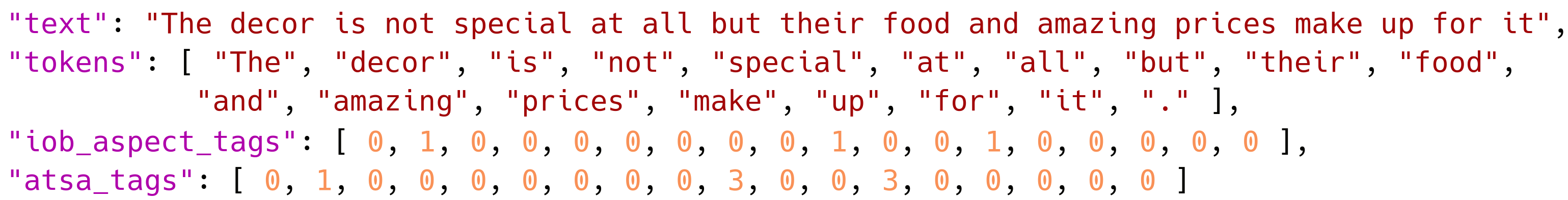}
\caption{Input data processed as JSON file}
\label{fig:inputdataasjson}
\end{figure}

Regardless of the transformer used, or the number of layers and networks added, the models are constructed so that each specific sub-task (ATE or ATSA) has a unique, generally available, input structure, in order to allow for easy scaling, upgrades and changes.
The models receive data under a JSON format specific to each task.
The common properties of the two input structures are represented by the full text review and a list of all tokens in the input data.
For the training phase, the ATE task (Figure~\ref{fig:inputdataasjson} - JSON tag \textit{iob\_aspect\_tags}) needs a third property, consisting of the IOB tagging for the entry, where the O tags are marked with 0, B tags with 1 and I tags with 0.
For the Sentiment Analysis task (Figure~\ref{fig:inputdataasjson} - JSON tag \textit{atsa\_tags}), takes as a third property, a list of polarities for each token, where 1 represents the negative polarity, 2 neutral, 3 positive while 0 marks the fact that no polarity was identified for the specific token -- in this case the token is not a term of interest, thus, the polarity is not important.

For the \textsc{BiLSTM} model, we use the classic implementation of LSTM~\cite{Hochreiter1997} with 256 units per layer and a dropout rate of 0.3.
The input size for the BiLSTM layer is equal to the hidden layer size of the transformer.
For the \textsc{CNN-BiLSTM} model, we use a 1-dimensional convolution (Conv1D) layer.
The input size of the Conv1D layer is equal to the hidden layer size of the transformer, the output size is equal to the sequence length, and the kernel size is set to 3.
The BiLSTM layer of the \textsc{CNN-BiLSTM} model is initialized with the same parameters as for the \textsc{BiLSTM} model.
We train each model for 2 epochs using the ADAM~\cite{Kingma2015} optimizer, a learning rate of $10^{-5}$ and the cross entropy as the loss function.
Each model is saved, after the training phase to be loaded by the web application server.

\section{Experimental Results}\label{sec:experimental_results}

In this section, we present the datasets used for experiments, the experimental setup, and the results obtained with our models.
We conclude this section with a discussion.

\subsection{Datasets}

We evaluate our solution on two datasets \textit{SemEval-2016 Task 5 - Restaurants (SE2016T5R)} and \textit{MAMS (Multi-Aspect Multi-sentiment) dataset}.
SE2016T5R consists of 350 reviews containing $\sim2\,000$ sentences with one aspect, multiple or no aspects.
MAMS contains 4\,297 sentences with at least two aspects with different sentiment polarities.

\paragraph{\textbf{SemEval 2016 Task 5 - Restaurants (SE2016T5R)}}
SE2016T5R was proposed as a shared task at SemEval 2016 as a challenge for solving the ABSA problem~\cite{Pontiki2016}.
The dataset contains a total of 722 aspect terms identified.
The most frequent term is \textit{NULL}, meaning that no aspect was found and a general sentence polarity is offered (627 times), then the following terms are identified: \textit{food} (216 times), \textit{place} (124 times), \textit{service} (121 times) and the rest occurring less than 50 times each.
When it comes to reviews length, the majority of the reviews have between 3 and 8 sentences, the rest being situated at less than 10 sentences / review.

The most frequent categories are \textit{FOOD}, \textit{SERVICE}, and \textit{RESTAURANT}.
Each category has multiple sub-categories that classify the aspect terms in a more relevant manner.
The dataset presents the category in direct connection with the aspect term, overlapping the start and end indices in the review's text.
We observe that most of the reviews are inclined into more "sentence-polarity" than "aspect-based polarity".
Only less than 200 sentences have 2 or more different types of polarities.

\paragraph{\textbf{Multi-Aspect Multi-Sentiment (MAMS)}}
MAMS~\cite{Jiang2019} tries to address the shortcoming of other datasets by introducing reviews that contain more than one polarity.
Thus, MAMS contains 4\,297 sentences with at least two opinions per review, going up to 8 for a small number of sentences.
In total, there are 11\,186 opinions.
The total number of aspect terms identified is 2\,586, the five most common being \textit{food}, \textit{menu}, \textit{waiter}, \textit{service}, and \textit{dinner}.
Compared to the SE2016T5R dataset, the difficulty of the MAMS dataset lays in the number of aspects present in every review.
For polarities, MAMS contains 5\,042 aspects with a neutral polarity (the most common polarity), 3\,380 positive, and 2\,764 negative.
We observe that, even though the dataset is not perfectly balanced, there is still a better ratio between the distribution of polarities.

\subsection{Experimental setup}

We fine-tuned the transformers and trained the models on a DGX Server with the following configuration: 4 Tesla V100-DGXS GPU with 32 GB VRAM, Intel Xeon E6-2698 CPU, and 264GB RAM.

We executed 10 runs per model, for each dataset.
We split the data at random using an 80\%-20\% ratio for training and testing, respectively.
We keep the distribution ratio of labels for each split.
Each training phase consists of 2 epochs.
Each run is executed using batches of size 4.
For each run, we compute the accuracy and micro and macro precision, recall, and F1-Score.
We also track the total execution time (training and testing) for each model separately.

\subsection{Ablation results}\label{sec:results}

We assess the performance of each branch composing the proposed Ensemble cell used in ATE and ATSA sub-tasks individually, training a model for each branch. 
We evaluate these models for both datasets, i.e., SE2016T5R and MAMS. 
In the following tables, we mark the best scores with the color yellow (the brighter the yellow, the higher the value).
All the tables present the mean and standard deviation for each considered metric after 10 runs using an 80\%-20\% train-test split that maintains the class ratio by using stratified k-folds.
We use 10\% of the training set for validation during the model training.

To evaluate each branch of the ATESA-B{\AE}RT, we build an individual branch model. 
We compare these models with ATESA-B{\AE}RT to determine how performance is improved.
When analyzing each individual branch model, we observe that the performance improvement between the pre-trained and fine-tuned variants of the transformer models is slightly increased.
Although the training of ATESA-B{\AE}RT takes more time than the training of the branch models, ATESA-B{\AE}RT outperforms each individual branch model when analyzing evaluation metrics, i.e., accuracy, micro precision, macro precision, etc.

The difference is slightly notable between the pre-trained and fine-tuned variants of the transformer models.
However, the models obtained state-of-the-art performance for both datasets.
The difference between the size of the two datasets is clearly notable, with execution time being almost 2 times bigger for some models on the MAMS dataset.
Despite the difference between the computed micro and macro averages due to the unbalanced number of labels, the model kept high values for all the evaluation scores.

For the ATE task, for both datasets (Table~\ref{tbl:results-ate-semeval} and~\ref{tbl:results-ate-mams}), \textsc{ATESA-B{\AE}RT} obtains an accuracy of 99.96\% on the SE2016T5R dataset and 99.95\% on the MAMS dataset.
For the sentiment analysis task (Tables~\ref{tbl:results-absa-semeval} and~\ref{tbl:results-absa-mams}), the \textsc{Linear} model with BERT obtains an accuracy of 99.74\% on SE2016T5R dataset and 99.87\% on the MAMS dataset.
We observe that the models that employ BART are outperformed by the corresponding models that use BERT.
The classification task consists in assigning a small number of labels in a relatively medium sized text, so using BART can result in an added overhead in memory and time resources for a small improvement.

\begin{table*}[!htbp]
\caption{ATE results on the SE2016T5R dataset (Note: the highlighted cells show the best results)\label{tbl:results-ate-semeval}}
\resizebox{\textwidth}{!}{%
\begin{tabular}{|c|c|c|cc|cc|cc|r|}
\hline
\rowcolor[HTML]{EFEFEF} 
\cellcolor[HTML]{EFEFEF} & \cellcolor[HTML]{EFEFEF} & \cellcolor[HTML]{EFEFEF} & \multicolumn{2}{c|}{\cellcolor[HTML]{EFEFEF}\textbf{Precision}} & \multicolumn{2}{c|}{\cellcolor[HTML]{EFEFEF}\textbf{Recall}} & \multicolumn{2}{c|}{\cellcolor[HTML]{EFEFEF}\textbf{F1-Score}} & \cellcolor[HTML]{EFEFEF} \\ \cline{4-9}
\rowcolor[HTML]{EFEFEF} 
\multirow{-2}{*}{\cellcolor[HTML]{EFEFEF}\textbf{Model}} & \multirow{-2}{*}{\cellcolor[HTML]{EFEFEF}\textbf{Transformer}} & \multirow{-2}{*}{\cellcolor[HTML]{EFEFEF}\textbf{Accuracy}} & \multicolumn{1}{c|}{\cellcolor[HTML]{EFEFEF}Micro} & Macro & \multicolumn{1}{c|}{\cellcolor[HTML]{EFEFEF}Micro} & Macro & \multicolumn{1}{c|}{\cellcolor[HTML]{EFEFEF}Micro} & Macro & \multirow{-2}{*}{\cellcolor[HTML]{EFEFEF}\textbf{\begin{tabular}[c]{@{}c@{}}Execution\\ time (s)\end{tabular}}} \\ \hline

\textsc{Linear} & BERT pre-trained & 99.73 $\pm$ 0.03 & \multicolumn{1}{c|}{99.73 $\pm$ 0.03} & 76.16 $\pm$ 2.36 & \multicolumn{1}{c|}{99.73 $\pm$ 0.03} & 69.48 $\pm$ 4.28 & \multicolumn{1}{c|}{99.73 $\pm$ 0.03} & 69.71 $\pm$ 4.24 & \resultmidclr  166.24 $\pm$ 0.29  \\ \hline

\textsc{BiLSTM} & BERT pre-trained & 99.57 $\pm$ 0.05 & \multicolumn{1}{c|}{99.57 $\pm$ 0.05} & 57.67 $\pm$ 16.27 & \multicolumn{1}{c|}{99.57 $\pm$ 0.05} & 47.08 $\pm$ 8.62 & \multicolumn{1}{c|}{99.57 $\pm$ 0.05} & 47.92 $\pm$ 7.88 & 186.35 $\pm$ 2.57  \\ \hline

\textsc{CNN-BiLSTM} & BERT pre-trained & 99.50 $\pm$ 0.01 & \multicolumn{1}{c|}{99.50 $\pm$ 0.01} & 52.26 $\pm$ 8.96 & \multicolumn{1}{c|}{99.50 $\pm$ 0.01} & 35.66 $\pm$ 0.78 & \multicolumn{1}{c|}{99.50 $\pm$ 0.01} & 37.21 $\pm$ 1.16 & 185.98 $\pm$ 11.41 \\ \hline

\textsc{Linear} & BERT fine-tuned & \resulthighclr 99.86 $\pm$ 0.03 & \multicolumn{1}{c|}{\resulthighclr 99.86 $\pm$ 0.03} & \resulthighclr 87.32 $\pm$ 4.11 & \multicolumn{1}{c|}{\resulthighclr 99.86 $\pm$ 0.03} & \resultlowclr 87.32 $\pm$ 2.77 & \multicolumn{1}{c|}{\resulthighclr 99.86 $\pm$ 0.03} & \resultmidclr 87.11 $\pm$ 3.03 & \resulthighclr 163.39 $\pm$ 0.62 \\ \hline

\textsc{BiLSTM} & BERT fine-tuned & \resultlowclr \resultlowclr 99.78 $\pm$ 0.03 & \multicolumn{1}{c|}{\resultlowclr 99.78 $\pm$ 0.03} & 82.24 $\pm$ 1.83 & \multicolumn{1}{c|}{99.78 $\pm$ 0.03} & 72.81 $\pm$ 4.70 & \multicolumn{1}{c|}{\resultlowclr 99.78 $\pm$ 0.03} & 72.04 $\pm$ 5.80 & \resultlowclr 172.93 $\pm$ 8.33 \\ \hline

\textsc{CNN-BiLSTM} & BERT fine-tuned & 99.51 $\pm$ 0.03 & \multicolumn{1}{c|}{99.51 $\pm$ 0.03} & 52.30 $\pm$ 10.06 & \multicolumn{1}{c|}{99.51 $\pm$ 0.03} & 36.61 $\pm$ 3.13 & \multicolumn{1}{c|}{99.51 $\pm$ 0.03} & 38.23 $\pm$ 3.71 & 184.93 $\pm$ 2.04 \\ \hline

\textsc{Linear} & BART pre-trained & 99.70 $\pm$ 0.03 & \multicolumn{1}{c|}{99.70 $\pm$ 0.03} & 80.64 $\pm$ 2.68 & \multicolumn{1}{c|}{99.70 $\pm$ 0.03} & 83.51 $\pm$ 2.39 & \multicolumn{1}{c|}{99.70 $\pm$ 0.03} & 81.80 $\pm$ 1.78 &  211.28 $\pm$ 0.50 \\ \hline

\textsc{BiLSTM} & BART pre-trained & 99.57 $\pm$ 0.03 & \multicolumn{1}{c|}{99.57 $\pm$ 0.03} & 75.50 $\pm$ 2.69 & \multicolumn{1}{c|}{99.57 $\pm$ 0.03} & 67.89 $\pm$ 2.42 & \multicolumn{1}{c|}{99.57 $\pm$ 0.03} & 68.10 $\pm$ 4.46 & 243.29 $\pm$ 0.93 \\ \hline
 
\textsc{CNN-BiLSTM} & BART pre-trained & 99.33 $\pm$ 0.04 & \multicolumn{1}{c|}{99.33 $\pm$ 0.04} & 49.85 $\pm$ 10.50 & \multicolumn{1}{c|}{99.33 $\pm$ 0.03} & 35.68 $\pm$ 3.08 & \multicolumn{1}{c|}{99.33 $\pm$ 0.04} & 36.61 $\pm$ 3.89 & 243.40 $\pm$ 0.86 \\ \hline

\textsc{Linear} & BART fine-tuned & \resultmidclr 99.79 $\pm$ 0.02 & \multicolumn{1}{c|}{\resultmidclr 99.79 $\pm$ 0.02} & \resultmidclr 86.16 $\pm$ 1.71 & \multicolumn{1}{c|}{\resultmidclr 99.79 $\pm$ 0.02} & \resulthighclr 92.24 $\pm$ 1.10 & \multicolumn{1}{c|}{\resultmidclr 99.79 $\pm$ 0.02} & \resulthighclr 88.94 $\pm$ 0.83 & 213.07 $\pm$ 13.81 \\ \hline

\textsc{BiLSTM} & BART fine-tuned & 99.77 $\pm$ 0.03 & \multicolumn{1}{c|}{99.77 $\pm$ 0.03} & \resultlowclr 84.64 $\pm$ 2.48 & \multicolumn{1}{c|}{99.77 $\pm$ 0.03} & \resultmidclr 89.38 $\pm$ 0.92 & \multicolumn{1}{c|}{99.77 $\pm$ 0.03} & \resultlowclr 86.75 $\pm$ 1.50 & 246.07 $\pm$ 20.36 \\ \hline

\textsc{CNN-BiLSTM} & BART fine-tuned & 99.34 $\pm$ 0.03 & \multicolumn{1}{c|}{99.34 $\pm$ 0.03} & 47.65 $\pm$ 12.05 & \multicolumn{1}{c|}{99.34 $\pm$ 0.03} & 34.78 $\pm$ 1.61 & \multicolumn{1}{c|}{99.34 $\pm$ 0.03} & 35.77 $\pm$ 2.77 & 245.83 $\pm$ 12.57 \\ \hline \hline

\multicolumn{2}{|c|}{\textsc{ATESA-B{\AE}RT}}  & \resulthighclr \textbf{99.96 $\pm$ 0.01} & \multicolumn{1}{c|}{\resulthighclr \textbf{99.96 $\pm$ 0.01}} & \resulthighclr \textbf{92.52 $\pm$ 0.01} & \multicolumn{1}{c|}{\resulthighclr \textbf{99.96 $\pm$ 0.01}} & \resulthighclr \textbf{95.12 $\pm$ 0.3} & \multicolumn{1}{c|}{\resulthighclr\textbf{ 99.96 $\pm$ 0.01}} & \resulthighclr \textbf{93.80 $\pm$ 0.01} & 258.39 $\pm$ 13.82 \\ \hline

\end{tabular}
}
\end{table*}

\begin{table*}[!htbp]
\caption{ATE results on the MAMS dataset (Note: the highlighted cells show the best results)\label{tbl:results-ate-mams}}
\resizebox{\textwidth}{!}{%
\begin{tabular}{|c|c|c|cc|cc|cc|r|}
\hline
\rowcolor[HTML]{EFEFEF} 
\cellcolor[HTML]{EFEFEF} & \cellcolor[HTML]{EFEFEF} & \cellcolor[HTML]{EFEFEF} & \multicolumn{2}{c|}{\cellcolor[HTML]{EFEFEF}\textbf{Precision}} & \multicolumn{2}{c|}{\cellcolor[HTML]{EFEFEF}\textbf{Recall}} & \multicolumn{2}{c|}{\cellcolor[HTML]{EFEFEF}\textbf{F1-Score}} & \cellcolor[HTML]{EFEFEF} \\ \cline{4-9}
\rowcolor[HTML]{EFEFEF} 
\multirow{-2}{*}{\cellcolor[HTML]{EFEFEF}\textbf{Model}} & \multirow{-2}{*}{\cellcolor[HTML]{EFEFEF}\textbf{Transformer}} & \multirow{-2}{*}{\cellcolor[HTML]{EFEFEF}\textbf{Accuracy}} & \multicolumn{1}{c|}{\cellcolor[HTML]{EFEFEF}Micro} & Macro & \multicolumn{1}{c|}{\cellcolor[HTML]{EFEFEF}Micro} & Macro & \multicolumn{1}{c|}{\cellcolor[HTML]{EFEFEF}Micro} & Macro & \multirow{-2}{*}{\cellcolor[HTML]{EFEFEF}\textbf{\begin{tabular}[c]{@{}c@{}}Execution\\ time (s)\end{tabular}}} \\ \hline

\textsc{Linear} & BERT pre-trained & \resultlowclr 99.78 $\pm$ 0.02 & \multicolumn{1}{c|}{\resultlowclr 99.78 $\pm$ 0.02} & \resultlowclr 89.17 $\pm$ 1.42 & \multicolumn{1}{c|}{\resultlowclr 99.78 $\pm$ 0.02} & 94.13 $\pm$ 2.08 & \multicolumn{1}{c|}{\resultlowclr 99.78 $\pm$ 0.02} & \resultlowclr 91.47 $\pm$ 0.97 & \resulthighclr 716.07 $\pm$ 3.96\\ \hline

\textsc{BiLSTM} & BERT pre-trained & 99.74 $\pm$ 0.03 & \multicolumn{1}{c|}{99.74 $\pm$ 0.03} & 85.44 $\pm$ 1.78 & \multicolumn{1}{c|}{99.74 $\pm$ 0.03} & 94.90 $\pm$ 1.13 & \multicolumn{1}{c|}{99.74 $\pm$ 0.03} & 89.72 $\pm$ 0.92 & 797.36 $\pm$ 1.38\\ \hline

\textsc{CNN-BiLSTM} & BERT pre-trained & 99.20 $\pm$ 0.06 & \multicolumn{1}{c|}{99.20 $\pm$ 0.06} & 70.70 $\pm$ 1.55 & \multicolumn{1}{c|}{99.20 $\pm$ 0.06} & 51.87 $\pm$ 7.30 & \multicolumn{1}{c|}{99.20 $\pm$ 0.06} & 56.76 $\pm$ 8.28 & 808.32 $\pm$ 1.00\\ \hline

\textsc{Linear} & BERT fine-tuned & \resulthighclr 99.93 $\pm$ 0.01 & \multicolumn{1}{c|}{\resulthighclr 99.93 $\pm$ 0.01} & \resulthighclr 96.07 $\pm$ 0.62 & \multicolumn{1}{c|}{\resulthighclr 99.93 $\pm$ 0.01} & \resulthighclr 98.74 $\pm$ 0.37 & \multicolumn{1}{c|}{\resulthighclr 99.93 $\pm$ 0.01} & \resulthighclr 97.37 $\pm$ 0.20 & \resultmidclr 719.63 $\pm$ 11.26\\ \hline

\textsc{BiLSTM} & BERT fine-tuned & \resultmidclr 99.90 $\pm$ 0.01 & \multicolumn{1}{c|}{\resultmidclr 99.90 $\pm$ 0.01} & \resultmidclr 94.63 $\pm$ 0.81 & \multicolumn{1}{c|}{\resultmidclr 99.90 $\pm$ 0.01} & \resultmidclr 98.58 $\pm$ 0.53 & \multicolumn{1}{c|}{\resultmidclr 99.90 $\pm$ 0.01} & \resultmidclr 96.54 $\pm$ 0.38 & \resultlowclr 791.57 $\pm$ 24.18\\ \hline

\textsc{CNN-BiLSTM} & BERT fine-tuned & 99.40 $\pm$ 0.02 & \multicolumn{1}{c|}{99.40 $\pm$ 0.02} & 75.72 $\pm$ 1.57 & \multicolumn{1}{c|}{99.40 $\pm$ 0.02} & 71.82 $\pm$ 2.77 & \multicolumn{1}{c|}{99.40 $\pm$ 0.02} & 73.16 $\pm$ 1.65 & 816.20 $\pm$ 4.86\\ \hline

\textsc{Linear} & BART pre-trained & 99.50 $\pm$ 0.04 & \multicolumn{1}{c|}{99.50 $\pm$ 0.04} & 80.94 $\pm$ 2.06 & \multicolumn{1}{c|}{99.50 $\pm$ 0.04} & 91.05 $\pm$ 1.19 & \multicolumn{1}{c|}{99.50 $\pm$ 0.04} & 85.30 $\pm$ 1.15 &  959.79 $\pm$ 2.34 \\ \hline

\textsc{BiLSTM} & BART pre-trained & 99.48 $\pm$ 0.03 & \multicolumn{1}{c|}{99.48 $\pm$ 0.03} & 79.48 $\pm$ 1.64 & \multicolumn{1}{c|}{99.48 $\pm$ 0.03} & 92.40 $\pm$ 0.90 & \multicolumn{1}{c|}{99.48 $\pm$ 0.03} & 85.01 $\pm$ 0.86 & 1\,089.97 $\pm$ 6.72 \\ \hline

\textsc{CNN-BiLSTM} & BART pre-trained & 98.87 $\pm$ 0.06 & \multicolumn{1}{c|}{98.87 $\pm$ 0.06} & 65.12 $\pm$ 5.74 & \multicolumn{1}{c|}{98.87 $\pm$ 0.06} & 57.16 $\pm$ 6.51 & \multicolumn{1}{c|}{98.87 $\pm$ 0.06} & 60.00 $\pm$ 5.90 & 1\,104.66 $\pm$ 21.54 \\ \hline

\textsc{Linear} & BART fine-tuned & 99.68 $\pm$ 0.02 & \multicolumn{1}{c|}{99.68 $\pm$ 0.02} & 86.95 $\pm$ 1.16 & \multicolumn{1}{c|}{99.68 $\pm$ 0.02} & \resultlowclr 96.20 $\pm$ 0.41 & \multicolumn{1}{c|}{99.68 $\pm$ 0.02} & 91.12 $\pm$ 0.83 &  947.81 $\pm$ 3.50 \\ \hline

\textsc{BiLSTM} & BART fine-tuned & 99.68 $\pm$ 0.02 & \multicolumn{1}{c|}{99.68 $\pm$ 0.02} & 87.14 $\pm$ 0.88 & \multicolumn{1}{c|}{99.68 $\pm$ 0.02} & 96.14 $\pm$ 0.37 & \multicolumn{1}{c|}{99.68 $\pm$ 0.02} & 91.27 $\pm$ 0.51 & 1\,090.74 $\pm$ 2.53 \\ \hline

\textsc{CNN-BiLSTM} & BART fine-tuned & 99.06 $\pm$ 0.07 & \multicolumn{1}{c|}{99.06 $\pm$ 0.07} & 70.82 $\pm$ 2.69 & \multicolumn{1}{c|}{99.06 $\pm$ 0.07} & 66.28 $\pm$ 4.40 & \multicolumn{1}{c|}{99.06 $\pm$ 0.07} & 67.96 $\pm$ 3.08 &  1\,088.94 $\pm$ 2.01 \\ \hline \hline

\multicolumn{2}{|c|}{\textsc{ATESA-B{\AE}RT}}  & \resulthighclr \textbf{99.95 $\pm$ 0.01} & \multicolumn{1}{c|}{\resulthighclr \textbf{99.95 $\pm$ 0.01}} & \resulthighclr \textbf{99.12 $\pm$ 0.01} & \multicolumn{1}{c|}{\resulthighclr \textbf{99.95 $\pm$ 0.01}} & \resulthighclr \textbf{98.32 $\pm$ 0.1} & \multicolumn{1}{c|}{\resulthighclr\textbf{ 99.53 $\pm$ 0.01}} & \resulthighclr \textbf{92.38 $\pm$ 0.03} & 1\.094.84 $\pm$ 1.82  \\ \hline

\end{tabular}
}
\end{table*}

\begin{table*}[!htbp]
\caption{ATSA results on the SE2016T5R dataset (Note: the highlighted cells show the best results)\label{tbl:results-absa-semeval}}
\resizebox{\textwidth}{!}{%
\begin{tabular}{|c|c|c|cc|cc|cc|c|}
\hline
\rowcolor[HTML]{EFEFEF} 
\cellcolor[HTML]{EFEFEF} & \cellcolor[HTML]{EFEFEF} & \cellcolor[HTML]{EFEFEF} & \multicolumn{2}{c|}{\cellcolor[HTML]{EFEFEF}\textbf{Precision}} & \multicolumn{2}{c|}{\cellcolor[HTML]{EFEFEF}\textbf{Recall}} & \multicolumn{2}{c|}{\cellcolor[HTML]{EFEFEF}\textbf{F1-Score}} & \cellcolor[HTML]{EFEFEF} \\ \cline{4-9}
\rowcolor[HTML]{EFEFEF} 
\multirow{-2}{*}{\cellcolor[HTML]{EFEFEF}\textbf{Model}} & \multirow{-2}{*}{\cellcolor[HTML]{EFEFEF}\textbf{Transformer}} & \multirow{-2}{*}{\cellcolor[HTML]{EFEFEF}\textbf{Accuracy}} & \multicolumn{1}{c|}{\cellcolor[HTML]{EFEFEF}Micro} & Macro & \multicolumn{1}{c|}{\cellcolor[HTML]{EFEFEF}Micro} & Macro & \multicolumn{1}{c|}{\cellcolor[HTML]{EFEFEF}Micro} & Macro & \multirow{-2}{*}{\cellcolor[HTML]{EFEFEF}\textbf{\begin{tabular}[c]{@{}c@{}}Execution\\ time (s)\end{tabular}}} \\ \hline

\textsc{Linear} & BERT pre-trained & \resultlowclr 99.73 $\pm$ 0.02 & \multicolumn{1}{c|}{\resultlowclr 99.73 $\pm$ 0.02 } & 42.79 $\pm$ 6.40 & \multicolumn{1}{c|}{\resultlowclr 99.73 $\pm$ 0.02 } & 46.31 $\pm$ 1.18 & \multicolumn{1}{c|}{\resultlowclr 99.73 $\pm$ 0.02 } & 43.29 $\pm$ 0.96 & \resultmidclr 169.98 $\pm$ 5.81\\ \hline

\textsc{BiLSTM} & BERT pre-trained & 99.58 $\pm$ 0.04 & \multicolumn{1}{c|}{99.58 $\pm$ 0.04} & 42.67 $\pm$ 2.39 & \multicolumn{1}{c|}{99.58 $\pm$ 0.04} & 34.03 $\pm$ 5.19 & \multicolumn{1}{c|}{99.58 $\pm$ 0.04} & 35.77 $\pm$ 4.31 & 183.75 $\pm$ 4.04\\ \hline

\textsc{CNN-BiLSTM} & BERT pre-trained & 99.49 $\pm$ 0.01 & \multicolumn{1}{c|}{99.49 $\pm$ 0.01} & 35.95 $\pm$ 10.56 & \multicolumn{1}{c|}{99.49 $\pm$ 0.01} & 25.13 $\pm$ 0.26 & \multicolumn{1}{c|}{99.49 $\pm$ 0.01} & 25.19 $\pm$ 0.50 & 181.44 $\pm$ 9.71\\ \hline

\textsc{Linear} & BERT fine-tuned & \resulthighclr 99.79 $\pm$ 0.01 & \multicolumn{1}{c|}{\resulthighclr 99.79 $\pm$ 0.01} & \resultmidclr 56.93 $\pm$ 10.99 & \multicolumn{1}{c|}{\resulthighclr 99.79 $\pm$ 0.01} & \resultmidclr 49.23 $\pm$ 2.46 & \multicolumn{1}{c|}{\resulthighclr 99.79 $\pm$ 0.01} & \resultmidclr 46.82 $\pm$ 3.75 & \resulthighclr 163.59 $\pm$ 2.94\\ \hline

\textsc{BiLSTM} & BERT fine-tuned & \resultmidclr 99.74 $\pm$ 0.02 & \multicolumn{1}{c|}{\resultmidclr 99.74 $\pm$ 0.02} & 41.09 $\pm$ 0.83 & \multicolumn{1}{c|}{\resultmidclr 99.74 $\pm$ 0.02} & 46.94 $\pm$ 0.33 & \multicolumn{1}{c|}{\resultmidclr 99.74 $\pm$ 0.02} & 43.55 $\pm$ 0.53 & \resultlowclr   178.91 $\pm$ 9.38\\ \hline

\textsc{CNN-BiLSTM} & BERT fine-tuned & 99.51 $\pm$ 0.02 & \multicolumn{1}{c|}{99.51 $\pm$ 0.02} & 36.31 $\pm$ 12.46 & \multicolumn{1}{c|}{99.51 $\pm$ 0.02} & 25.44 $\pm$ 0.81 & \multicolumn{1}{c|}{99.51 $\pm$ 0.02} & 25.74 $\pm$ 1.44 & 190.83 $\pm$ 4.83\\ \hline

\textsc{Linear} & BART pre-trained & 99.62 $\pm$ 0.02 & \multicolumn{1}{c|}{99.62 $\pm$ 0.02} & \resultlowclr 49.95 $\pm$ 4.28 & \multicolumn{1}{c|}{99.62 $\pm$ 0.02} & 46.92 $\pm$ 0.80 & \multicolumn{1}{c|}{99.62 $\pm$ 0.02} & \resultlowclr 44.44 $\pm$ 1.43 & 213.05 $\pm$ 1.42 \\ \hline

\textsc{BiLSTM} & BART pre-trained & 99.53 $\pm$ 0.03 & \multicolumn{1}{c|}{99.53 $\pm$ 0.03} & 39.54 $\pm$ 0.68 & \multicolumn{1}{c|}{99.53 $\pm$ 0.03} & 41.23 $\pm$ 2.79 & \multicolumn{1}{c|}{99.53 $\pm$ 0.03} & 40.20 $\pm$ 1.04 & 243.48 $\pm$ 0.81 \\ \hline

\textsc{CNN-BiLSTM} & BART pre-trained & 99.34 $\pm$ 0.02 & \multicolumn{1}{c|}{99.34 $\pm$ 0.02} & 30.06 $\pm$ 7.98 & \multicolumn{1}{c|}{99.34 $\pm$ 0.02} & 25.11 $\pm$ 0.18 & \multicolumn{1}{c|}{99.34 $\pm$ 0.02} & 25.13 $\pm$ 0.35 & 242.74 $\pm$ 0.81 \\ \hline

\textsc{Linear} & BART fine-tuned & \resultlowclr 99.73 $\pm$ 0.03 & \multicolumn{1}{c|}{\resultlowclr 99.73 $\pm$ 0.03} & \resulthighclr 58.69 $\pm$ 2.68 & \multicolumn{1}{c|}{\resultlowclr 99.73 $\pm$ 0.03} & \resulthighclr 55.47 $\pm$ 4.88 & \multicolumn{1}{c|}{\resultlowclr 99.73 $\pm$ 0.03} & \resulthighclr 54.50 $\pm$ 5.02 & 208.67 $\pm$ 1.38 \\ \hline

\textsc{BiLSTM} & BART fine-tuned & 99.68 $\pm$ 0.02 & \multicolumn{1}{c|}{99.68 $\pm$ 0.02} & 41.00 $\pm$ 0.75 & \multicolumn{1}{c|}{99.68 $\pm$ 0.02} & \resultlowclr 47.60 $\pm$ 0.75 & \multicolumn{1}{c|}{99.68 $\pm$ 0.02} & 43.72 $\pm$ 0.42 & 239.87 $\pm$ 0.44 \\ \hline

\textsc{CNN-BiLSTM} & BART fine-tuned & 99.36 $\pm$ 0.03 & \multicolumn{1}{c|}{99.36 $\pm$ 0.03} & 33.15 $\pm$ 7.39 & \multicolumn{1}{c|}{99.36 $\pm$ 0.03} & 26.82 $\pm$ 3.20 & \multicolumn{1}{c|}{99.36 $\pm$ 0.03} & 27.31 $\pm$ 3.82 & 241.19 $\pm$ 0.90 \\ \hline \hline

\multicolumn{2}{|c|}{\textsc{ATESA-B{\AE}RT}}  & \resulthighclr \textbf{99.84 $\pm$ 0.01} & \multicolumn{1}{c|}{\resulthighclr \textbf{99.84 $\pm$ 0.01}} & \resulthighclr \textbf{92.71 $\pm$ 0.07} & \multicolumn{1}{c|}{\resulthighclr \textbf{99.84 $\pm$ 0.01}} & \resulthighclr \textbf{91.43 $\pm$ 0.1} & \multicolumn{1}{c|}{\resulthighclr\textbf{ 99.84 $\pm$ 0.01}} & \resulthighclr \textbf{ 92.07 $\pm$ 0.01} & 242.72 $\pm$ 0.75 \\ \hline

\end{tabular}
}
\end{table*}

\begin{table*}[!htbp]
\caption{ATSA results on the MAMS dataset (Note: the highlighted cells show the best results)\label{tbl:results-absa-mams}}
\resizebox{\textwidth}{!}{%
\begin{tabular}{|c|c|c|cc|cc|cc|r|}
\hline
\rowcolor[HTML]{EFEFEF} 
\cellcolor[HTML]{EFEFEF} & \cellcolor[HTML]{EFEFEF} & \cellcolor[HTML]{EFEFEF} & \multicolumn{2}{c|}{\cellcolor[HTML]{EFEFEF}\textbf{Precision}} & \multicolumn{2}{c|}{\cellcolor[HTML]{EFEFEF}\textbf{Recall}} & \multicolumn{2}{c|}{\cellcolor[HTML]{EFEFEF}\textbf{F1-Score}} & \cellcolor[HTML]{EFEFEF} \\ \cline{4-9}
\rowcolor[HTML]{EFEFEF} 
\multirow{-2}{*}{\cellcolor[HTML]{EFEFEF}\textbf{Model}} & \multirow{-2}{*}{\cellcolor[HTML]{EFEFEF}\textbf{Transformer}} & \multirow{-2}{*}{\cellcolor[HTML]{EFEFEF}\textbf{Accuracy}} & \multicolumn{1}{c|}{\cellcolor[HTML]{EFEFEF}Micro} & Macro & \multicolumn{1}{c|}{\cellcolor[HTML]{EFEFEF}Micro} & Macro & \multicolumn{1}{c|}{\cellcolor[HTML]{EFEFEF}Micro} & Macro & \multirow{-2}{*}{\cellcolor[HTML]{EFEFEF}\textbf{\begin{tabular}[c]{@{}c@{}}Execution\\ time (s)\end{tabular}}} \\ \hline

\textsc{Linear} & BERT pre-trained & \resultlowclr 99.69 $\pm$ 0.02 & \multicolumn{1}{c|}{\resultlowclr 99.69 $\pm$ 0.02 } & 82.68 $\pm$ 1.55 & \multicolumn{1}{c|}{\resultlowclr 99.69 $\pm$ 0.02 } & 88.20 $\pm$ 2.17 & \multicolumn{1}{c|}{\resultlowclr 99.69 $\pm$ 0.02 } & 85.20 $\pm$ 0.73 & \resultmidclr  723.00 $\pm$ 6.40\\ \hline

\textsc{BiLSTM} & BERT pre-trained & 99.62 $\pm$ 0.03 & \multicolumn{1}{c|}{99.62 $\pm$ 0.03} & 78.08 $\pm$ 1.78 & \multicolumn{1}{c|}{99.62 $\pm$ 0.03} & 85.84 $\pm$ 1.39 & \multicolumn{1}{c|}{99.62 $\pm$ 0.03} & 81.58 $\pm$ 1.21 & 797.49 $\pm$ 0.75\\ \hline

\textsc{CNN-BiLSTM} & BERT pre-trained & 99.12 $\pm$ 0.03 & \multicolumn{1}{c|}{99.12 $\pm$ 0.03} & 49.65 $\pm$ 3.42 & \multicolumn{1}{c|}{99.12 $\pm$ 0.03} & 31.52 $\pm$ 3.39 & \multicolumn{1}{c|}{99.12 $\pm$ 0.03} & 33.56 $\pm$ 3.98 & 806.10 $\pm$ 4.05\\ \hline

\textsc{Linear} & BERT fine-tuned & \resulthighclr 99.87 $\pm$ 0.01 & \multicolumn{1}{c|}{\resulthighclr 99.87 $\pm$ 0.01} & \resulthighclr 91.44 $\pm$ 1.10 & \multicolumn{1}{c|}{\resulthighclr 99.87 $\pm$ 0.01} & \resulthighclr 96.04 $\pm$ 0.48 & \multicolumn{1}{c|}{\resulthighclr 99.87 $\pm$ 0.01} & \resulthighclr 93.64 $\pm$ 0.46 & \resulthighclr  718.52 $\pm$ 0.60\\ \hline

\textsc{BiLSTM} & BERT fine-tuned & \resultmidclr 99.84 $\pm$ 0.01 & \multicolumn{1}{c|}{\resultmidclr 99.84 $\pm$ 0.01} & \resultmidclr 89.42 $\pm$ 1.36 & \multicolumn{1}{c|}{\resultmidclr 99.84 $\pm$ 0.01} & \resultmidclr 95.29 $\pm$ 0.73 & \multicolumn{1}{c|}{\resultmidclr 99.84 $\pm$ 0.01} & \resultmidclr 92.18 $\pm$ 0.63 & \resultlowclr 792.16 $\pm$ 0.51\\ \hline

\textsc{CNN-BiLSTM} & BERT fine-tuned & 99.19 $\pm$ 0.02 & \multicolumn{1}{c|}{99.19 $\pm$ 0.02} & 49.09 $\pm$ 8.84 & \multicolumn{1}{c|}{99.19 $\pm$ 0.02} & 40.42 $\pm$ 1.79 & \multicolumn{1}{c|}{99.19 $\pm$0.02 } & 40.75 $\pm$ 2.01 & 806.11 $\pm$ 0.93\\ \hline

\textsc{Linear} & BART pre-trained & 99.41 $\pm$ 0.04 & \multicolumn{1}{c|}{99.41 $\pm$ 0.04} & 76.74 $\pm$ 2.12 & \multicolumn{1}{c|}{99.41 $\pm$ 0.04} & 84.11 $\pm$ 1.34 & \multicolumn{1}{c|}{99.41 $\pm$ 0.04} & 80.04 $\pm$ 1.03 & 948.75 $\pm$ 1.42 \\ \hline

\textsc{BiLSTM} & BART pre-trained & 99.35 $\pm$ 0.02 & \multicolumn{1}{c|}{99.35 $\pm$ 0.02} & 73.30 $\pm$ 0.90 & \multicolumn{1}{c|}{99.35 $\pm$ 0.02} & 82.69 $\pm$ 1.54 & \multicolumn{1}{c|}{99.35 $\pm$ 0.02} & 77.36 $\pm$ 0.81 & 1083.69 $\pm$ 0.95 \\ \hline

\textsc{CNN-BiLSTM} & BART pre-trained & 98.75 $\pm$ 0.04 & \multicolumn{1}{c|}{98.75 $\pm$ 0.04} & 49.41 $\pm$ 6.37 & \multicolumn{1}{c|}{98.75 $\pm$ 0.04} & 33.97 $\pm$ 4.65 & \multicolumn{1}{c|}{98.75 $\pm$ 0.04} & 36.44 $\pm$ 5.42 & 1\,088.83 $\pm$ 5.80 \\ \hline

\textsc{Linear} & BART fine-tuned & 99.61 $\pm$ 0.03 & \multicolumn{1}{c|}{99.61 $\pm$ 0.03} & \resultlowclr 82.87 $\pm$ 1.57 & \multicolumn{1}{c|}{99.61 $\pm$ 0.03} & 91.91 $\pm$ 0.43 & \multicolumn{1}{c|}{99.61 $\pm$ 0.03} &  87.00 $\pm$ 0.86 & 944.19 $\pm$ 2.13 \\ \hline

\textsc{BiLSTM} & BART fine-tuned & 99.58 $\pm$ 0.05 & \multicolumn{1}{c|}{99.58 $\pm$ 0.05} & 81.65 $\pm$ 2.25 & \multicolumn{1}{c|}{99.58 $\pm$ 0.05} & \resultlowclr 92.16 $\pm$ 0.56 & \multicolumn{1}{c|}{99.58 $\pm$ 0.05} & 86.39 $\pm$ 1.37 & 1\,093.80 $\pm$ 3.60 \\ \hline

\textsc{CNN-BiLSTM} & BART fine-tuned & 98.86 $\pm$ 0.04 & \multicolumn{1}{c|}{98.86 $\pm$ 0.04} & 55.91 $\pm$ 3.71 & \multicolumn{1}{c|}{98.86 $\pm$ 0.04} & 42.25 $\pm$ 3.69 & \multicolumn{1}{c|}{98.86 $\pm$ 0.04} & 45.26 $\pm$ 4.57 & 1\,086.15 $\pm$ 2.69 \\ \hline \hline

\multicolumn{2}{|c|}{\textsc{ATESA-B{\AE}RT}}  & \resulthighclr \textbf{99.93 $\pm$ 0.01} & \multicolumn{1}{c|}{\resulthighclr \textbf{99.93 $\pm$ 0.01}} & \resulthighclr \textbf{93.62 $\pm$ 0.01} & \multicolumn{1}{c|}{\resulthighclr \textbf{99.93 $\pm$ 0.01}} & \resulthighclr \textbf{97.72 $\pm$ 0.1} & \multicolumn{1}{c|}{\resulthighclr\textbf{ 99.93 $\pm$ 0.01}} & \resulthighclr \textbf{92.38 $\pm$ 0.01} & 1\.098.24 $\pm$ 2.71 \\ \hline

\end{tabular}
}
\end{table*}

Figure~\ref{fig:performance} presents the runtime comparison for each task on each dataset w.r.t. the used transformer model.
The models managed to maintain a similar impact over the average execution time for both tasks w.r.t. the dataset.
We observe that the models that use BERT are faster than the ones that use BART.
On MAMS, both tasks need longer execution times than SE2016T5R due to the following factors: 
(1) a larger dataset, and 
(2) more opinions per sentence.

\pgfplotsset{height=0.20\textwidth, width=0.48\columnwidth,
/pgfplots/ybar legend/.style={
        /pgfplots/legend image code/.code={
        \draw[##1,/tikz/.cd,bar width=3pt,yshift=-0.2em,bar shift=0pt]
                plot coordinates {(0cm,0.8em)};},
                every axis/.append style={font=\tiny},
},
}
\begin{figure}[!ht]
    \centering
    \subfloat[ATE task on SE2016T5R dataset]{%
    \begin{tikzpicture}[]
        \begin{axis}[
        xmin=0,
        xmax=4,
        ymin=0,
        ymax=350,
        bar width=5pt,
        ybar = 2pt,
        xtick \empty, 
        extra x ticks={1,2,3}, 
        extra x tick labels={\textsc{Linear}, \textsc{BiLSTM}, \textsc{CNN-BiLSTM}},
        ylabel={Seconds},
        legend style={anchor=north, legend columns=4, legend cell align=left, at={(0.5,1.32)}},
        ]

        \addplot+[color=blue, pattern=north east lines, pattern color=blue] [error bars/.cd, y dir = both, y explicit] table [x=MODEL, y=BERT_PT_AVG, y error = BERT_PT_STD, col sep=comma] {ATE_SE2016T5R.csv};
        \addlegendentry{BERT Pre-Trained};
        
        \addplot+[color=red, pattern=north west lines, pattern color=red] [error bars/.cd, y dir = both, y explicit] table [x=MODEL, y=BERT_FT_AVG, y error = BERT_FT_STD, col sep=comma] {ATE_SE2016T5R.csv};
        \addlegendentry{BERT Fine-Tuned};
        
        \addplot+[color=green, pattern=horizontal lines, pattern color=green] [error bars/.cd, y dir = both, y explicit] table [x=MODEL, y=BART_PT_AVG, y error = BART_PT_STD, col sep=comma] {ATE_SE2016T5R.csv};
        \addlegendentry{BART Pre-Trained};
        
        \addplot+[color=black, pattern=north east lines, pattern color=black] [error bars/.cd, y dir = both, y explicit] table [x=MODEL, y=BART_FT_AVG, y error = BART_FT_STD, col sep=comma] {ATE_SE2016T5R.csv};
        \addlegendentry{BART Fine-Tuned};
    \end{axis}
    \end{tikzpicture}
    \label{fig:ate_se20t5r_tp}
    }%
    \hfill%
    \subfloat[ATSA task on SE2016T5R dataset]{%
    \begin{tikzpicture}[]
        \begin{axis}[
        xmin=0,
        xmax=4,
        ymin=0,
        ymax=350,
        bar width=5pt,
        ybar = 2pt,
        xtick \empty, 
        extra x ticks={1,2,3}, 
        extra x tick labels={\textsc{Linear}, \textsc{BiLSTM}, \textsc{CNN-BiLSTM}},
        ylabel={Seconds},
        legend style={anchor=north, legend columns=4, legend cell align=left, at={(0.5,1.32)}},
        ]

        \addplot+[color=blue, pattern=north east lines, pattern color=blue] [error bars/.cd, y dir = both, y explicit] table [x=MODEL, y=BERT_PT_AVG, y error = BERT_PT_STD, col sep=comma] {ATSA_SE2016T5R.csv};
        \addlegendentry{BERT Pre-Trained};
        
        \addplot+[color=red, pattern=north west lines, pattern color=red] [error bars/.cd, y dir = both, y explicit] table [x=MODEL, y=BERT_FT_AVG, y error = BERT_FT_STD, col sep=comma] {ATSA_SE2016T5R.csv};
        \addlegendentry{BERT Fine-Tuned};
        
        \addplot+[color=green, pattern=horizontal lines, pattern color=green] [error bars/.cd, y dir = both, y explicit] table [x=MODEL, y=BART_PT_AVG, y error = BART_PT_STD, col sep=comma] {ATSA_SE2016T5R.csv};
        \addlegendentry{BART Pre-Trained};
        
        \addplot+[color=black, pattern=north east lines, pattern color=black] [error bars/.cd, y dir = both, y explicit] table [x=MODEL, y=BART_FT_AVG, y error = BART_FT_STD, col sep=comma] {ATSA_SE2016T5R.csv};
        \addlegendentry{BART Fine-Tuned};
    \end{axis}
    \end{tikzpicture}
    \label{fig:atsa_se20t5r_tp}   
    }%
    \hfill%
    \subfloat[ATE task on MAMS dataset]{%
    \begin{tikzpicture}[]
        \begin{axis}[
        xmin=0,
        xmax=4,
        ymin=0,
        ymax=1200,
        bar width=5pt,
        ybar = 2pt,
        xtick \empty, 
        extra x ticks={1,2,3}, 
        extra x tick labels={\textsc{Linear}, \textsc{BiLSTM}, \textsc{CNN-BiLSTM}},
        ylabel={Seconds},
legend style={anchor=north, legend columns=4, legend cell align=left, at={(0.5,1.32)}},
        ]

        \addplot+[color=blue, pattern=north east lines, pattern color=blue] [error bars/.cd, y dir = both, y explicit] table [x=MODEL, y=BERT_PT_AVG, y error = BERT_PT_STD, col sep=comma] {ATE_MAMS.csv};
        \addlegendentry{BERT Pre-Trained};
        
        \addplot+[color=red, pattern=north west lines, pattern color=red] [error bars/.cd, y dir = both, y explicit] table [x=MODEL, y=BERT_FT_AVG, y error = BERT_FT_STD, col sep=comma] {ATE_MAMS.csv};
        \addlegendentry{BERT Fine-Tuned};
        
        \addplot+[color=green, pattern=horizontal lines, pattern color=green] [error bars/.cd, y dir = both, y explicit] table [x=MODEL, y=BART_PT_AVG, y error = BART_PT_STD, col sep=comma] {ATE_MAMS.csv};
        \addlegendentry{BART Pre-Trained};
        
        \addplot+[color=black, pattern=north east lines, pattern color=black] [error bars/.cd, y dir = both, y explicit] table [x=MODEL, y=BART_FT_AVG, y error = BART_FT_STD, col sep=comma] {ATE_MAMS.csv};
        \addlegendentry{BART Fine-Tuned};
    \end{axis}
    \end{tikzpicture}
    \label{fig:ate_mams_tp}   
    }%
    \subfloat[ATSA task on MAMS dataset]{%
    \begin{tikzpicture}[]
        \begin{axis}[
        xmin=0,
        xmax=4,
        ymin=0,
        ymax=1200,
        bar width=5pt,
        ybar = 2pt,
        xtick \empty, 
        extra x ticks={1,2,3}, 
        extra x tick labels={\textsc{Linear}, \textsc{BiLSTM}, \textsc{CNN-BiLSTM}},
        ylabel={Seconds},
legend style={anchor=north, legend columns=4, legend cell align=left, at={(0.5,1.32)}},
        ]

        \addplot+[color=blue, pattern=north east lines, pattern color=blue] [error bars/.cd, y dir = both, y explicit] table [x=MODEL, y=BERT_PT_AVG, y error = BERT_PT_STD, col sep=comma] {ATSA_MAMS.csv};
        \addlegendentry{BERT Pre-Trained};
        
        \addplot+[color=red, pattern=north west lines, pattern color=red] [error bars/.cd, y dir = both, y explicit] table [x=MODEL, y=BERT_FT_AVG, y error = BERT_FT_STD, col sep=comma] {ATSA_MAMS.csv};
        \addlegendentry{BERT Fine-Tuned};
        
        \addplot+[color=green, pattern=horizontal lines, pattern color=green] [error bars/.cd, y dir = both, y explicit] table [x=MODEL, y=BART_PT_AVG, y error = BART_PT_STD, col sep=comma] {ATSA_MAMS.csv};
        \addlegendentry{BART Pre-Trained};
        
        \addplot+[color=black, pattern=north east lines, pattern color=black] [error bars/.cd, y dir = both, y explicit] table [x=MODEL, y=BART_FT_AVG, y error = BART_FT_STD, col sep=comma] {ATSA_MAMS.csv};
        \addlegendentry{BART Fine-Tuned};
    \end{axis}
    \end{tikzpicture}
    \label{fig:atsa_mams_tp}   
    }
    \caption{Performance time comparison}
    \label{fig:performance}   
\end{figure}
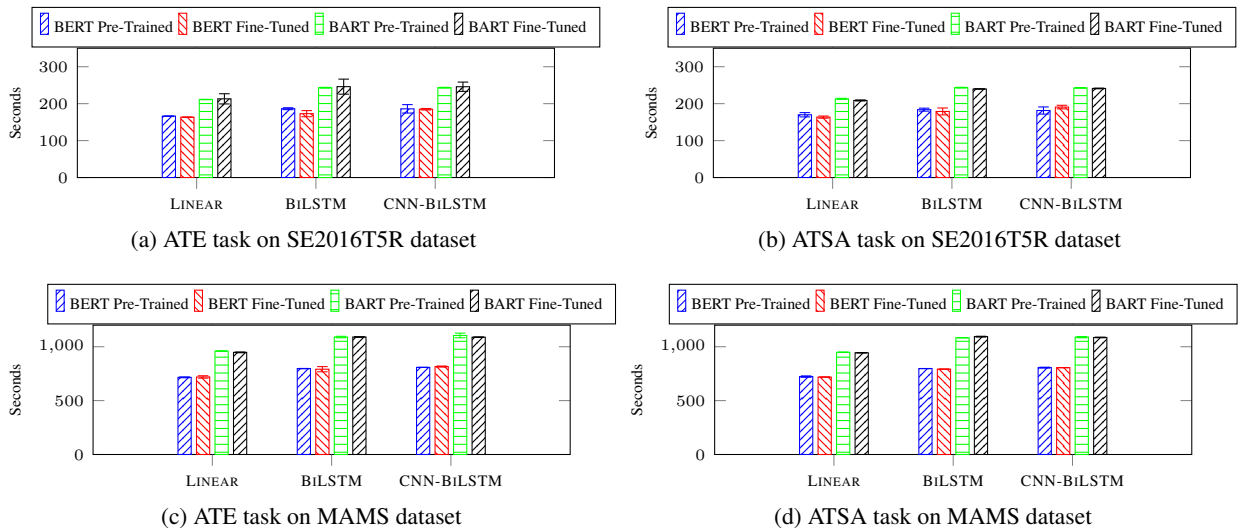

\subsection{Comparison results}

We compare the results obtained by ATESA-B{\AE}RT with previous results from the state of the art.
We observe that ATESA-B{\AE}RT outperforms the state-of-the-art models on both ATE and ATSA tasks.

For the ATE task, we compare the results obtained by the proposed ensemble with the ones obtained by models that use either SemEval 2014 Task 4 on the Restaurants (SE2014T4R) dataset or SE2016T5R.
The models we use for this comparison are: 
\begin{itemize}
\item \textbf{CRF}~\cite{Chernyshevich2014IHSRB} uses Conditional Random Fields together with IOB tagging and part-of-speech analysis for sequence labeling.

\item \textbf{W+L+D} and \textbf{W+L+D+B}~\cite{unsupervisedwordanddeppathforate} use the following main concepts: \textbf{W}ord embeddings, \textbf{L}inear context embedding features, \textbf{D}ependency context embedding features and \textbf{B}aseline feature templates.
    
\item \textbf{CMLA}~\cite{cmlamodelate} (Coupled Multi-Layer Attentions) is based on constructing different auxiliary pairs of attentions for ATE and Opinion Term Extraction that aim to assign an attention score for each token.

\item \textbf{DE-CNN-CRF} and \textbf{DE-CNN}~\cite{Xu2018} use IOB tagging and two embeddings, i.e., a general purpose embedding and a domain-specific embedding, that are concatenated and redirected into a CNN.
\end{itemize}

Table~\ref{tbl:comparisonate} presents the results for the ATE task.
ATESA-B{\AE}RT manages to outperform the state-of-the-art models, obtaining an F1-Score of 93.80\%, while \textbf{W+L+D+B}~\cite{unsupervisedwordanddeppathforate} obtains an F1-score of 84.97\% on SE2014T4R  and \textbf{DE-CNN}~\cite{Xu2018} obtains 74.37\% on SE2016T5R.

\begin{table}[!htbp]
\centering
\caption{
ATE results comparison on the SemEval datasets 
(Note: highlighted cells show the best results)
}
\label{tbl:comparisonate}
\begin{tabular}{|l|c|r|}
\hline
\rowcolor[HTML]{EFEFEF} 
\textbf{Model} & \textbf{Dataset} & \textbf{F1-Score} \\ \hline
CRF~\cite{Chernyshevich2014IHSRB}                                                        & SE2014T4R & 79.62 \\ \hline
W+L+D~\cite{unsupervisedwordanddeppathforate}                                                      & SE2014T4R & 84.31 \\ \hline
W+L+D+B~\cite{unsupervisedwordanddeppathforate}                                                   & SE2014T4R & 84.97 \\ \hline
CMLA~\cite{cmlamodelate}                                                       & SE2014T4R & 77.80 \\ \hline
DE-CNN-CRF~\cite{Xu2018}                                                 & SE2016T5R & 74.10 \\ \hline
DE-CNN~\cite{Xu2018}                                                     & SE2016T5R & 74.37 \\ \hline
\textbf{\textcolor{blue}{ATESA-B{\AE}RT}} - our solution & SE2016T5R & \resulthighclr \textbf{93.80} \\ \hline
\end{tabular}
\end{table}

For the ATSA task, we use the same two SemEval datasets, i.e., SE2014T4R and SE2016T5R, with the following models in our comparison:
\begin{itemize}
    \item \textbf{BERT-single}, \textbf{BERT-pair-QA-M}, \textbf{BERT-pair-NLI-M}, \textbf{BERT-pair-QA-B}, \textbf{BERT-pair-NLI-B}~\cite{Sun2019} use a sentence-pair classification method to infer the polarity;
    
    \item \textbf{BERT for ABSA}~\cite{absausingbert} uses a simple fine-tuned version of BERT for solving out-of-domain aspect classification;
    
    \item \textbf{DeBERTa}~\cite{Marcacini2021} employs a disentangled attention mechanism for simplifying and extracting the semantic and syntactic features from BERT, using a fine-tuned variant of the transformer.
\end{itemize}

Table~\ref{tbl:comparisonsa} presents the results for the ATSA task.
We observe that our model ATESA-B{\AE}RT, with an accuracy of $99.84\%$, outperforms the state-of-the-art models.
From the state-of-the-art models, the BERT-pair-QA-B obtains an accuracy of $95.6\%$, with $\sim 4\%$ less than our ensemble model.
 
\begin{table}[!htbp]
\centering
\caption{ATSA comparison on the SemEval dataset
(Note: highlighted cells show the best results)
}\label{tbl:comparisonsa}
\begin{tabular}{|l|c|r|}
\hline
\cellcolor[HTML]{EFEFEF}\textbf{Model} & \cellcolor[HTML]{EFEFEF}\textbf{Dataset} & \cellcolor[HTML]{EFEFEF}\textbf{Accuracy} \\ \hline
BERT-single~\cite{Sun2019}                                                 & SE2014T4R & 93.3 \\ \hline
BERT-pair-QA-M~\cite{Sun2019}                                              & SE2014T4R & 95.4 \\ \hline
BERT-pair-NLI-M~\cite{Sun2019}                                             & SE2014T4R & 94.4 \\ \hline
BERT-pair-QA-B~\cite{Sun2019}                                              & SE2014T4R & 95.6 \\ \hline
BERT-pair-NLI-B~\cite{Sun2019}                                             & SE2014T4R & 95.1 \\ \hline
BERT for ABSA~\cite{absausingbert}                                               & SE2016T5R & 89.8 \\ \hline
DeBERTa~\cite{Marcacini2021}                                                     & SE2016T5R & 89.46 \\ \hline
\textbf{\textcolor{blue}{ATESA-B{\AE}RT}} - our solution  & SE2016T5R & \resulthighclr \textbf{99.84} \\ \hline

\end{tabular}
\end{table}

For the MAMS dataset, we compare ATESA-B{\AE}RT results with the ones obtained for the ATSA task by the following state-of-the-art-models:
\begin{itemize}
    \item \textbf{HAGNN-BERT} and \textbf{HAGNN-GloVe}~\cite{An2022} are heterogeneous graph neural networks that encode words, aspects, and sentences as nodes.;
    
    \item \textbf{CapsNet-BERT}~\cite{Jiang2019} uses Capsule Networks and BERT to model and interpret the contextual connections for the aspects;
    
    \item \textbf{CapsNet-BERT-DR}~\cite{Jiang2019} is a \textbf{CapsNet-BERT} extension that uses a dynamic routing mechanism~\cite{McGill2017} to inference the context;
    
    \item \textbf{RoBERTa-TMM}~\cite{Wu2020} captures the direct dependencies between words and sentences as well as the polarity of each aspect term;
    
    \item \textbf{TransEncAsp+SCAPT} and \textbf{BERTAsp+SCAPT}~\cite{li-etal-2021-learning-implicit} are two context-aware solutions used to determine the aspect term polarity;
    
    \item \textbf{RGAT-BERT}~\cite{Bai_2021} is a relational graph attention network with syntactic dependency information that incorporates label features into the attention mechanism.
\end{itemize}

Table~\ref{tbl:atsacomparisonmams} presents the results obtained on MAMS for the ATSA task.
The overall best model with an accuracy of $99.93\%$ is ATESA-B{\AE}RT.
The best state-of-the-art model, i.e., RoBERTa-TMM, obtains an accuracy of $85.64\%$, with $\sim 14\%$ less than our ensemble model.

\begin{table}[!htbp]
\centering
\caption{ATSA results comparison on the MAMS dataset
(Note: highlighted cells show the best results)
}
\label{tbl:atsacomparisonmams}
\begin{tabular}{|l|r|}
    \hline
    \cellcolor[HTML]{EFEFEF}\textbf{Model} &  \cellcolor[HTML]{EFEFEF}\textbf{Accuracy} \\ \hline
    HAGNN-BERT~\cite{An2022}        & 66.92 \\ \hline
    HAGNN-GloVe~\cite{An2022}       & 72.58 \\ \hline
    CapsNet-BERT~\cite{Jiang2019}      & 83.39 \\ \hline
    CapsNet-BERT-DR~\cite{Jiang2019}   & 82.97 \\ \hline
    RoBERTa-TMM~\cite{Wu2020}       & 85.64 \\ \hline
    TransEncAsp+SCAPT~\cite{li-etal-2021-learning-implicit} & 80.54 \\ \hline
    BERTAsp+SCAPT~\cite{li-etal-2021-learning-implicit}     & 85.63 \\ \hline
    RGAT-BERT~\cite{Bai_2021}         & 84.52 \\ \hline
    \textbf{\textcolor{blue}{ATESA-B{\AE}RT}} - our solution  & \resulthighclr \textbf{99.93} \\ \hline
\end{tabular}
\end{table}

\subsection{Discussions}

Aspect-based sentiment analysis is a complex problem, and even the most performant models may have difficulties analyzing sentences with many aspects.
We observed that in previous research, the most common datasets used are the ones proposed at SemEval yearly.
One of the main problems we find with these datasets is that their distributions of the positive, negative, and neutral labels are highly unbalanced.
To showcase the efficiency of ATESA-B{\AE}RT, our ensemble model,for the task of ABSA, we perform extensive tests on the SemEval dataset for a general comparison with other model and on the MAMS dataset in order to observe the behavior on a larger dataset that contains only multi-aspect and multi-sentiment pairs.

Contextual representation of textual data for accurately determining and extracting aspect terms is one of the main issues that needs to be addressed.
Since both BERT and BART embeddings are contextual, we ensure that our ensemble model is going to identify with high accuracy the dependencies between the parts of speech and sentences in order to extract the targeted aspect terms.
Moreover, by fine-tuning both transformers and using IOB tagging, we manage to further improve the accuracy of the ATE task.
To make sure that our ensemble model performs well when the input sentence has more than two aspects, we use multiple neural network models that underline the contextual dependencies, i.e., \textsc{BiLSTM} and \textsc{BiLSTM-CNN}.

The fine-tuning procedures of BERT and BART models specialized the models in the restaurant domain through iterating and analyzing both corpora, i.e., SE2016T5R and MAMS.
In order to switch the target of the models to a different domain, the fine-tuning procedure has to be re-done on new domain-targeted datasets.
Specializing a model on a specific dataset increases the evaluation metrics scores over the learned aspects and sentence structures, with a higher miss detection rate on new unlearned input.
Furthermore, the experimental evaluations show that, by fine-tuning both BERT and BART, we manage to obtain high accuracy scores with a small number of training epochs on small and medium sized datasets.
We also note that, when taken individually, the models that use BART did not manage to outperform BERT, although BART has additional resource requirements with over 30 million more trainable parameters than BERT.

In the end, we can not confirm that the ABSA problem has been fully solved on general inputs, due to a lack of public datasets that contain multi-domain reviews, but we can claim with high confidence that, even with a small to medium sized dataset containing multi-aspect, multi-polarity pairs reviews from a specific domain, we can obtain high accuracy results.

\section{Conclusions}\label{sec:conclusions}

Sentiment Analysis is one of the main tools used to determine the opinion of used users from product and service reviews.
Extracting the aspects from online reviews with corresponding polarities can result in high efficiency in understanding customer problems for different domains, ranging from simple shops and restaurants to online retailers and business-to-business scenarios.

In this paper, we propose, design and implement a new heterogeneous ensemble architecture, ATESA-B{\AE}RT, consisting of 12 different models that use both pre-trained and fine-tuned BERT and BART transformers, in order to solve the Aspect-Based Sentiment Analysis problem.
We split the problem into two sub-problems, for a better understanding and to better understand and observe the behavior of the models: 
(1) Aspect Term Extraction (ATE), and 
(2) Aspect Term Sentiment Analysis (ATSA).
We test ATESA-B{\AE}RT on two datasets: textit{SemEval16 Task 5 Restaurants} (SE2016T5R)~\cite{Pontiki2016} and \textit{Multi Aspect Multi Sentiment} (MAMS)~\cite{Jiang2019}.
The experimental results show that the proposed model obtains a high accuracy after training it for only 2 epochs using a train-test split of 80\%-20\%.

ATESA-B{\AE}RT obtains an accuracy of 99.96\% on MAMS and 99.95\% on SE2016T5R for the ATE task.
For the ATSA task, the model obtains an accuracy of 99.84\% for the SE2016T5R dataset and 99.93\% for MAMS.

In conclusion, we managed to obtain high evaluation metrics scores results for the ABSA task over the restaurant domain with an ensemble model that uses 12 different neural networks over two different datasets, for both sub-problems of ATE and ATSA.

As a future direction, the current solution we aim to analyze our model on new datasets that combines multiple corpora from different domains, in order to understand if the ABSA problem can be generalized and solved through a single, cross-domain, implementation.

\bibliographystyle{plainnat}  
\bibliography{main}

\begin{thebibliography}{40}
\providecommand{\natexlab}[1]{#1}
\providecommand{\url}[1]{\texttt{#1}}
\expandafter\ifx\csname urlstyle\endcsname\relax
  \providecommand{\doi}[1]{doi: #1}\else
  \providecommand{\doi}{doi: \begingroup \urlstyle{rm}\Url}\fi

\bibitem[An et~al.(2022)An, Tian, Chen, and Zheng]{An2022}
Wenbin An, Feng Tian, Ping Chen, and Qinghua Zheng.
\newblock Aspect-based sentiment analysis with heterogeneous graph neural
  network.
\newblock \emph{{IEEE} Transactions on Computational Social Systems}, pages
  1--10, 2022.
\newblock \doi{10.1109/tcss.2022.3148866}.

\bibitem[Bai et~al.(2021)Bai, Liu, and Zhang]{Bai_2021}
Xuefeng Bai, Pengbo Liu, and Yue Zhang.
\newblock {Investigating Typed Syntactic Dependencies for Targeted Sentiment
  Classification Using Graph Attention Neural Network}.
\newblock \emph{{IEEE}/{ACM} Transactions on Audio, Speech, and Language
  Processing}, 29:\penalty0 503--514, 2021.
\newblock \doi{10.1109/taslp.2020.3042009}.

\bibitem[Bensoltane and Zaki(2021)]{Bensoltane2021}
Rajae Bensoltane and Taher Zaki.
\newblock Towards arabic aspect-based sentiment analysis: a transfer
  learning-based approach.
\newblock \emph{Social Network Analysis and Mining}, 12\penalty0 (1), nov 2021.
\newblock \doi{10.1007/s13278-021-00794-4}.

\bibitem[Chernyshevich(2014)]{Chernyshevich2014IHSRB}
Maryna Chernyshevich.
\newblock {IHS} {R}{\&}{D} {B}elarus: Cross-domain extraction of product
  features using {CRF}.
\newblock In \emph{International Workshop on Semantic Evaluation}, pages
  309--313, 2014.
\newblock \doi{10.3115/v1/S14-2051}.

\bibitem[Cho et~al.(2013)Cho, Okazaki, Miwa, and Tsujii]{Cho2013}
Han-Cheol Cho, Naoaki Okazaki, Makoto Miwa, and Jun'Ichi Tsujii.
\newblock Named entity recognition with multiple segment representations.
\newblock \emph{Information Processing \& Management}, 49\penalty0
  (4):\penalty0 954–--965, 2013.
\newblock ISSN 0306-4573.
\newblock \doi{10.1016/j.ipm.2013.03.002}.

\bibitem[Devlin et~al.(2019)Devlin, Chang, Lee, and Toutanova]{Devlin2019}
Jacob Devlin, Ming-Wei Chang, Kenton Lee, and Kristina Toutanova.
\newblock Bert: Pre-training of deep bidirectional transformers for language
  understanding.
\newblock In \emph{Conference of the North American Chapter of the Association
  for Computational Linguistics}, pages 4171--4186, 2019.
\newblock \doi{10.18653/v1/N19-1423}.

\bibitem[Do et~al.(2019)Do, Prasad, Maag, and Alsadoon]{Do2019}
Hai~Ha Do, Penatiyana Withanage~Chandana Prasad, Angelika Maag, and Abeer
  Alsadoon.
\newblock Deep learning for aspect-based sentiment analysis: A comparative
  review.
\newblock \emph{Expert Systems with Applications}, 118:\penalty0 272--299,
  2019.
\newblock ISSN 0957-4174.
\newblock \doi{10.1016/j.eswa.2018.10.003}.

\bibitem[He et~al.(2021)He, Liu, Gao, and Chen]{PengchengHe2021}
Pengcheng He, Xiaodong Liu, Jianfeng Gao, and Weizhu Chen.
\newblock Deberta: decoding-enhanced bert with disentangled attention.
\newblock In \emph{International Conference on Learning Representations ({ICLR}
  2021)}, 2021.

\bibitem[He et~al.(2017)He, Lee, Ng, and Dahlmeier]{he-etal-2017-unsupervised}
Ruidan He, Wee~Sun Lee, Hwee~Tou Ng, and Daniel Dahlmeier.
\newblock An unsupervised neural attention model for aspect extraction.
\newblock In \emph{Annual Meeting of the Association for Computational
  Linguistics}, pages 388--397, 2017.
\newblock URL \url{https://aclanthology.org/P17-1036}.

\bibitem[Hoang et~al.(2019)Hoang, Bihorac, and Rouces]{absausingbert}
Mickel Hoang, Oskar~Alija Bihorac, and Jacobo Rouces.
\newblock Aspect-based sentiment analysis using {BERT}.
\newblock In \emph{Nordic Conference on Computational Linguistics}, pages
  187--196, 09 2019.

\bibitem[Hochreiter and Schmidhuber(1997)]{Hochreiter1997}
Sepp Hochreiter and Jürgen Schmidhuber.
\newblock Long short-term memory.
\newblock \emph{Neural computation}, 9:\penalty0 1735--80, 12 1997.
\newblock \doi{10.1162/neco.1997.9.8.1735}.

\bibitem[Honnibal et~al.(2020)Honnibal, Montani, Van~Landeghem, and
  Boyd]{SpaCy}
Matthew Honnibal, Ines Montani, Sofie Van~Landeghem, and Adriane Boyd.
\newblock {spaCy: Industrial-strength Natural Language Processing in Python},
  2020.
\newblock URL \url{https://spacy.io/}.

\bibitem[Jiang et~al.(2019)Jiang, Chen, Xu, Ao, and Yang]{Jiang2019}
Qingnan Jiang, Lei Chen, Ruifeng Xu, Xiang Ao, and Min Yang.
\newblock A challenge dataset and effective models for aspect-based sentiment
  analysis.
\newblock In \emph{Conference on Empirical Methods in Natural Language
  Processing and International Joint Conference on Natural Language Processing
  (EMNLP-IJCNLP)}, pages 6280--6285, 2019.
\newblock \doi{10.18653/v1/D19-1654}.

\bibitem[Kingma and Ba(2015)]{Kingma2015}
Diederik~P Kingma and Jimmy Ba.
\newblock Adam: A method for stochastic optimization.
\newblock In \emph{The International Conference on Learning Representations
  (ICLR)}, pages 1--15, 2015.

\bibitem[Lewis et~al.(2020)Lewis, Liu, Goyal, Ghazvininejad, Mohamed, Levy,
  Stoyanov, and Zettlemoyer]{Lewis2020}
Mike Lewis, Yinhan Liu, Naman Goyal, Marjan Ghazvininejad, Abdelrahman Mohamed,
  Omer Levy, Veselin Stoyanov, and Luke Zettlemoyer.
\newblock {BART}: Denoising sequence-to-sequence pre-training for natural
  language generation, translation, and comprehension.
\newblock In \emph{Annual Meeting of the Association for Computational
  Linguistics}, pages 7871--7880, 2020.
\newblock \doi{10.18653/v1/2020.acl-main.703}.

\bibitem[Li et~al.(2021{\natexlab{a}})Li, Chen, Feng, Ma, Wang, and
  Hovy]{Li2021}
Ruifan Li, Hao Chen, Fangxiang Feng, Zhanyu Ma, Xiaojie Wang, and Eduard Hovy.
\newblock Dual graph convolutional networks for aspect-based sentiment
  analysis.
\newblock In \emph{Annual Meeting of the Association for Computational
  Linguistics and International Joint Conference on Natural Language Processing
  (ACl-IJNLP)}, pages 6319--6329, 2021{\natexlab{a}}.
\newblock \doi{10.18653/v1/2021.acl-long.494}.

\bibitem[Li et~al.(2021{\natexlab{b}})Li, Zou, Zhang, Zhang, and
  Wei]{li-etal-2021-learning-implicit}
Zhengyan Li, Yicheng Zou, Chong Zhang, Qi~Zhang, and Zhongyu Wei.
\newblock {Learning Implicit Sentiment in Aspect-based Sentiment Analysis with
  Supervised Contrastive Pre-Training}.
\newblock In \emph{Conference on Empirical Methods in Natural Language
  Processing}, pages 246--256, 2021{\natexlab{b}}.
\newblock \doi{10.18653/v1/2021.emnlp-main.22}.

\bibitem[Liang et~al.(2022)Liang, Su, Gui, Cambria, and Xu]{Liang2022}
Bin Liang, Hang Su, Lin Gui, Erik Cambria, and Ruifeng Xu.
\newblock Aspect-based sentiment analysis via affective knowledge enhanced
  graph convolutional networks.
\newblock \emph{Knowledge-Based Systems}, 235:\penalty0 107643, 2022.
\newblock ISSN 0950-7051.
\newblock \doi{10.1016/j.knosys.2021.107643}.

\bibitem[Liu et~al.(2019)Liu, Ott, Goyal, Du, Joshi, Chen, Levy, Lewis,
  Zettlemoyer, and Stoyanov]{Liu2019}
Yinhan Liu, Myle Ott, Naman Goyal, Jingfei Du, Mandar Joshi, Danqi Chen, Omer
  Levy, Mike Lewis, Luke Zettlemoyer, and Veselin Stoyanov.
\newblock Roberta: A robustly optimized bert pretraining approach, 2019.

\bibitem[Marcacini and Silva(2021)]{Marcacini2021}
Ricardo~Marcondes Marcacini and Emanuel Silva.
\newblock {Aspect-based Sentiment Analysis using BERT with Disentangled
  Attention}.
\newblock In \emph{LatinX in AI at International Conference on Machine Learning
  2021}, pages 1--4, 2021.

\bibitem[McGill and Perona(2017)]{McGill2017}
Mason McGill and Pietro Perona.
\newblock Deciding how to decide: Dynamic routing in artificial neural
  networks.
\newblock In Doina Precup and Yee~Whye Teh, editors, \emph{International
  Conference on Machine Learning}, pages 2363--2372, 2017.

\bibitem[Mitroi et~al.(2020)Mitroi, Truica, Apostol, and Florea]{Mitroi2020}
Madalina Mitroi, Ciprian-Octavian Truica, Elena-Simona Apostol, and Adina~Magda
  Florea.
\newblock Sentiment analysis using topic-document embeddings.
\newblock In \emph{International Conference on Intelligent Computer
  Communication and Processing}, pages 75--82, sep 2020.
\newblock \doi{10.1109/iccp51029.2020.9266181}.

\bibitem[Paszke et~al.(2019)Paszke, Gross, Massa, Lerer, Bradbury, Chanan,
  Killeen, Lin, Gimelshein, Antiga, Desmaison, Kopf, Yang, DeVito, Raison,
  Tejani, Chilamkurthy, Steiner, Fang, Bai, and Chintala]{Paszke2019}
Adam Paszke, Sam Gross, Francisco Massa, Adam Lerer, James Bradbury, Gregory
  Chanan, Trevor Killeen, Zeming Lin, Natalia Gimelshein, Luca Antiga, Alban
  Desmaison, Andreas Kopf, Edward Yang, Zachary DeVito, Martin Raison, Alykhan
  Tejani, Sasank Chilamkurthy, Benoit Steiner, Lu~Fang, Junjie Bai, and Soumith
  Chintala.
\newblock Pytorch: An imperative style, high-performance deep learning library.
\newblock In H.~Wallach, H.~Larochelle, A.~Beygelzimer, F.~d\textquotesingle
  Alch\'{e}-Buc, E.~Fox, and R.~Garnett, editors, \emph{Advances in Neural
  Information Processing Systems}, volume~32. Curran Associates, Inc., 2019.
\newblock URL
  \url{https://proceedings.neurips.cc/paper/2019/file/bdbca288fee7f92f2bfa9f7012727740-Paper.pdf}.

\bibitem[Petrescu et~al.(2019)Petrescu, Truica, and Apostol]{Petrescu2019}
Alexandru Petrescu, Ciprian-Octavian Truica, and Elena-Simona Apostol.
\newblock Sentiment analysis of events in social media.
\newblock In \emph{International Conference on Intelligent Computer
  Communication and Processing}, pages 143--149, sep 2019.
\newblock \doi{10.1109/iccp48234.2019.8959677}.

\bibitem[Petrescu et~al.(2023)Petrescu, Truică, Apostol, and
  Paschke]{Petrescu2023}
Alexandru Petrescu, Ciprian-Octavian Truică, Elena-Simona Apostol, and Adrian
  Paschke.
\newblock {EDSA-Ensemble: an Event Detection Sentiment Analysis Ensemble
  Architecture}, 2023.

\bibitem[Phan and Ogunbona(2020)]{Phan2020}
Minh~Hieu Phan and Philip~O. Ogunbona.
\newblock Modelling context and syntactical features for aspect-based sentiment
  analysis.
\newblock In \emph{Annual Meeting of the Association for Computational
  Linguistics}, pages 3211--3220, July 2020.
\newblock \doi{10.18653/v1/2020.acl-main.293}.

\bibitem[Pontiki et~al.(2016)Pontiki, Galanis, Papageorgiou, Androutsopoulos,
  Manandhar, AL-Smadi, Al-Ayyoub, Zhao, Qin, de~clercq, Hoste, Apidianaki,
  Tannier, Loukachevitch, Kotelnikov, Bel, Zafra, and Eryiğit]{Pontiki2016}
Maria Pontiki, Dimitris Galanis, Haris Papageorgiou, Ion Androutsopoulos,
  Suresh Manandhar, Mohammad AL-Smadi, Mahmoud Al-Ayyoub, Yanyan Zhao, Bing
  Qin, Orphee de~clercq, Véronique Hoste, Marianna Apidianaki, Xavier Tannier,
  Natalia Loukachevitch, Evgeny Kotelnikov, Nuria Bel, Salud~María Zafra, and
  Gülşen Eryiğit.
\newblock Semeval-2016 task 5: Aspect based sentiment analysis.
\newblock In \emph{International Workshop on Semantic Evaluation}, pages
  19--30, 2016.
\newblock \doi{10.18653/v1/S16-1002}.

\bibitem[Pradhan and Sharma(2021)]{Pradhan2021}
Rahul Pradhan and Dilip~Kumar Sharma.
\newblock A frequency-based approach to extract aspect for aspect-based
  sentiment analysis.
\newblock In \emph{Proceedings of Second International Conference on Computing,
  Communications, and Cyber-Security}, pages 499--510, 2021.

\bibitem[Radford et~al.(2018)Radford, Narasimhan, Salimans, and
  Sutskever]{Radford2018}
Alec Radford, Karthik Narasimhan, Tim Salimans, and Ilya Sutskever.
\newblock {Improving Language Understanding by Generative Pre-Training}, 2018.

\bibitem[Ramshaw and Marcus(1995)]{Ramshaw1995}
Lance Ramshaw and Mitch Marcus.
\newblock Text chunking using transformation-based learning.
\newblock In \emph{Workshop on Very Large Corpora}, pages 82--94, 1995.
\newblock URL \url{https://aclanthology.org/W95-0107}.

\bibitem[Sun et~al.(2019)Sun, Huang, and Qiu]{Sun2019}
Chi Sun, Luyao Huang, and Xipeng Qiu.
\newblock Utilizing {BERT} for aspect-based sentiment analysis via constructing
  auxiliary sentence.
\newblock In \emph{Conference of the North {A}merican Chapter of the
  Association for Computational Linguistics}, pages 380--385, June 2019.
\newblock \doi{10.18653/v1/N19-1035}.

\bibitem[Tang et~al.(2016)Tang, Qin, Feng, and Liu]{Tang2016}
Duyu Tang, Bing Qin, Xiaocheng Feng, and Ting Liu.
\newblock Effective lstms for target-dependent sentiment classification.
\newblock In Nicoletta Calzolari, Yuji Matsumoto, and Rashmi Prasad, editors,
  \emph{International Conference on Computational Linguistics}, pages
  3298--3307. {ACL}, 2016.
\newblock URL \url{https://aclanthology.org/C16-1311/}.

\bibitem[Tao and Fang(2020)]{Tao2020}
Jie Tao and Xing Fang.
\newblock Toward multi-label sentiment analysis: a transfer learning based
  approach.
\newblock \emph{Journal of Big Data}, 7\penalty0 (1), jan 2020.
\newblock \doi{10.1186/s40537-019-0278-0}.

\bibitem[Truic{\u{a}} et~al.(2021)Truic{\u{a}}, Apostol, Șerban, and
  Paschke]{Truica2021}
Ciprian-Octavian Truic{\u{a}}, Elena-Simona Apostol, Maria-Luiza Șerban, and
  Adrian Paschke.
\newblock Topic-based document-level sentiment analysis using contextual cues.
\newblock \emph{Mathematics}, 9\penalty0 (21):\penalty0 1--23(2722), oct 2021.
\newblock \doi{10.3390/math9212722}.

\bibitem[Wang et~al.(2017)Wang, Pan, Dahlmeier, and Xiao]{cmlamodelate}
Wenya Wang, Sinno~Jialin Pan, Daniel Dahlmeier, and Xiaokui Xiao.
\newblock Coupled multi-layer attentions for co-extraction of aspect and
  opinion terms.
\newblock In \emph{AAAI Conference on Artificial Intelligence}, pages
  3316–--3322, 2017.

\bibitem[Wu et~al.(2020)Wu, Ying, Dai, Huang, and Chen]{Wu2020}
Zhen Wu, Chengcan Ying, Xinyu Dai, Shujian Huang, and Jiajun Chen.
\newblock Transformer-based multi-aspect modeling for multi-aspect
  multi-sentiment analysis.
\newblock In Xiaodan Zhu, Min Zhang, Yu~Hong, and Ruifang He, editors,
  \emph{{CCF} International Conference}, pages 546--557, 2020.
\newblock \doi{10.1007/978-3-030-60457-8\_45}.

\bibitem[Xu et~al.(2018)Xu, Liu, Shu, and Yu]{Xu2018}
Hu~Xu, Bing Liu, Lei Shu, and Philip~S. Yu.
\newblock Double embeddings and {CNN}-based sequence labeling for aspect
  extraction.
\newblock In \emph{Annual Meeting of the Association for Computational
  Linguistics}, pages 592--598, 2018.
\newblock \doi{10.18653/v1/P18-2094}.

\bibitem[Yang et~al.(2019)Yang, Dai, Yang, Carbonell, Salakhutdinov, and
  Le]{Yang2019}
Zhilin Yang, Zihang Dai, Yiming Yang, Jaime Carbonell, Russlan Salakhutdinov,
  and Quoc~V. Le.
\newblock Xlnet: Generalized autoregressive pretraining for language
  understanding.
\newblock In \emph{Advances in Neural Information Processing Systems}, pages
  5753--5763, 2019.

\bibitem[Yin et~al.(2016)Yin, Wei, Dong, Xu, Zhang, and
  Zhou]{unsupervisedwordanddeppathforate}
Yichun Yin, Furu Wei, Li~Dong, Kaimeng Xu, Ming Zhang, and Ming Zhou.
\newblock Unsupervised word and dependency path embeddings for aspect term
  extraction.
\newblock In \emph{International Joint Conference on Artificial Intelligence},
  page 2979–2985, 2016.
\newblock ISBN 9781577357704.

\bibitem[Łukasz Augustyniak et~al.(2021)Łukasz Augustyniak, Kajdanowicz, and
  Kazienko]{Augustyniak2021}
Łukasz Augustyniak, Tomasz Kajdanowicz, and Przemysław Kazienko.
\newblock Comprehensive analysis of aspect term extraction methods using
  various text embeddings.
\newblock \emph{Computer Speech \& Language}, 69:\penalty0 101217, 2021.
\newblock \doi{10.1016/j.csl.2021.101217}.

\end{thebibliography}

\end{document}